\documentclass[10pt,twocolumn,letterpaper,table]{article}

\pdfoutput=1

\usepackage{cvpr}              

\usepackage{graphicx}
\usepackage{amsmath}
\usepackage{amssymb}
\usepackage{booktabs}
\usepackage{float}
\usepackage{enumerate}
\usepackage[all]{nowidow}
\usepackage{uoftcolors}

\usepackage{mathtools} 
\usepackage{booktabs} 
\usepackage{tikz} 
\usepackage{amsfonts}
\usepackage{multirow}
\usepackage{multicol}
\usepackage{color,colortbl}
\usepackage{xcolor}
\usepackage{xspace}
\usepackage{wrapfig}
\usepackage{subcaption}
\usepackage{bbm}
\usepackage{bm}
\usepackage{url}

\usepackage{abbreviations}

\definecolor{cvprblue}{rgb}{0.21,0.49,0.74}

\def\eg{e.g.,\ }               
\def\ie{i.e.,\ }               

\definecolor{LavenderBlue}{rgb}{0.7020,    0.8039,    0.8902}
\definecolor{Lightapricot}{rgb}{0.9961,    0.8510,    0.6510}
\definecolor{thirdtablecolor}{rgb}{0.8706,    0.7961,    0.8941}

\newcommand{\heading}[1]{\noindent\textbf{#1}}

\long\def\ignorethis#1{}

\definecolor{demphcolor}{RGB}{100,100,100}

\newcommand{\dotproduct}[2]{\langle#1, #2\rangle}

\newcommand{\absvalue}[1]{\lvert#1\rvert}

\newlength\pagetopmargin
\newlength\figcapmargin
\newlength\figmargin
\newlength\tablecapmargin
\newlength\tablemargin

\newcommand{\appref}[1]{\ref{#1}}

\usepackage{soul}

\def\videotokens{\bm{z}}
\def\noise{\bm{\epsilon}}
\def\velocity{\bm{v}}
\def\ditmodel{u_\theta}
\def\textembs{\bm{e}}
\def\generallabel{\bm{y}}

\def\startt{t^{\mathrm{start}}}
\def\endt{t^{\mathrm{end}}}
\def\midt{t^{\mathrm{mid}}}
\def\densecaption{c}
\def\densecaptionembs{\textembs^{\mathrm{\densecaption}}}
\def\numevents{N^{\mathrm{e}}}

\def\cutscenet{t^{\mathrm{cut}}}
\def\numcuts{N^{\mathrm{cut}}}
\def\cuttoken{\textembs^{\mathrm{cut}}}

\def\ropeangle{\theta}
\def\ropebase{\theta_{\mathrm{base}}}
\def\ropetextembs{\tilde{\textembs}}
\def\ropecuttoken{\tilde{\textembs}^{\mathrm{cut}}}

\newcommand{\appropto}{\mathrel{\vcenter{
  \offinterlineskip\halign{\hfil$##$\cr
    \propto\cr\noalign{\kern2pt}\sim\cr\noalign{\kern-2pt}}}}}

\def\holdout{HoldOut\xspace}
\def\jointmodel{Concat\xspace}
\def\imgcondmodel{AutoReg\xspace}

\newcommand{\algoNameFull}{MinT\xspace}

\usepackage[pagebackref,breaklinks,colorlinks,allcolors=cvprblue]{hyperref}

\usepackage[capitalize]{cleveref}
\crefname{section}{Sec.}{Secs.}
\Crefname{section}{Section}{Sections}
\Crefname{table}{Table}{Tables}
\crefname{table}{Tab.}{Tabs.}
\Crefname{figure}{Figure}{Figures}
\crefname{figure}{Fig.}{Figs.}

\def\paperID{2215} 
\def\confName{CVPR}
\def\confYear{2025}

\title{Mind the Time: Temporally-Controlled Multi-Event Video Generation}

\author{
Ziyi Wu$^{1,2,3}$, Aliaksandr Siarohin$^{1}$, Willi Menapace$^{1}$, Ivan Skorokhodov$^{1}$, \\
Yuwei Fang$^{1}$, Varnith Chordia$^{1}$, Igor Gilitschenski$^{2,3,*}$, Sergey Tulyakov$^{1,*}$ \\
$^1$Snap Research, $^2$University of Toronto, $^3$Vector Institute
}

\begin{document}

\twocolumn[{
\renewcommand\twocolumn[1][]{#1}
\maketitle
\vspace{-1.3cm}
\begin{center}
    \centering
    \captionsetup{type=figure}
    \includegraphics[width=\textwidth]{imgs/qual/teaser.pdf}
    \vspace*{-0.71cm}
    \captionof{figure}{
    \textbf{Time-controlled multi-event video generation with \algoNameFull}.
    Given a sequence of event text prompts and their desired start and end timestamps, \algoNameFull synthesizes smoothly connected events with consistent subjects and backgrounds.
    In addition, it can control the time span of each event flexibly.
    Here, we show the results of sequential gestures, daily activities, facial expressions, and cat movements.
    }
    \label{fig:teaser}
    \vspace*{-0.2cm}
\end{center}
}]

\maketitle

\begin{abstract}

\vspace{-3mm}
Real-world videos consist of sequences of events. Generating such sequences with precise temporal control is infeasible with existing video generators that rely on a single paragraph of text as input. When tasked with generating multiple events described using a single prompt, such methods often ignore some of the events or fail to arrange them in the correct order.
To address this limitation, we present \algoNameFull, a multi-event video generator with temporal control. Our key insight is to bind each event to a specific period in the generated video, which allows the model to focus on one event at a time. To enable time-aware interactions between event captions and video tokens, we design a time-based positional encoding method, dubbed ReRoPE. This encoding helps to guide the cross-attention operation.
By fine-tuning a pre-trained video diffusion transformer on temporally grounded data, our approach produces coherent videos with smoothly connected events.
For the first time in the literature, our model offers control over the timing of events in generated videos.
Extensive experiments demonstrate that \algoNameFull outperforms existing commercial and open-source models by a large margin.
Additional results and details are available at our \href{https://mint-video.github.io/}{project page}.

\end{abstract}    

\vspace{-5mm}
\section{Introduction}
\label{sec:intro}

%
%
Recent research in video diffusion models~\cite{VideoDiffusionModels} has achieved tremendous progress~\cite{SVD,AlignYourLatents,ImagenVideo,MetaMovieGen,VideoCrafter1,VideoCrafter2,AnimateDiff}.
These approaches typically rely on a single text prompt, and generate videos capturing only a single event.
In contrast, real-world videos often comprise sequences of events with rich dynamics.
Thus, achieving realism requires the ability to generate multiple events with fine-grained temporal control~\cite{MEVG,Gen-L-Video}.

\begin{figure*}[!t]
    \vspace{\pagetopmargin}
    \vspace{-3mm}
    \centering
    \includegraphics[width=1.0\linewidth]{imgs/qual/sota-model-failure-case.pdf}
    \vspace{\figcapmargin}
    \vspace{-5mm}
    \caption{
        \textbf{Multi-event video generation results from SOTA video generators and \algoNameFull.}
        We run two open-source models CogVideoX-5B~\cite{CogVideoX} and Mochi 1~\cite{Mochi}, and two commercial models Kling 1.5~\cite{KLING1_5} and Gen-3 Alpha~\cite{Gen3Alpha} to generate sequential events.
        All of them only generate a subset of events while ignoring the remaining ones.
        In contrast, \algoNameFull generates a natural video with all events smoothly connected.
        Please refer to Appendix~\appref{app:more-compare-with-sota} and our \href{https://mint-video.github.io/\#compare-with-sota}{project page} for more comparisons.
        Comparisons with Sora~\cite{Sora} can be found \href{https://mint-video.github.io/\#compare-with-sora}{here}.
    }
    \label{fig:sota-model-failure-case}
    \vspace{\figmargin}
    \vspace{-1mm}
\end{figure*}


%
%
A naive solution to multi-event video generation is to concatenate all event descriptions into a single, extended prompt, such as ``A man raises his arms, lowers them down, and then moves them left and right".
However, \cref{fig:sota-model-failure-case} shows that even state-of-the-art video models struggle to produce satisfactory results from such prompts.
Some recent works tackle this problem in an autoregressive way~\cite{Phenaki,MEVG}.
They generate each event individually with its own prompt, and condition the model on the last frame of the previous event to ensure consistency.
Yet, they often generate stagnated video frames with limited motion~\cite{StreamingT2V,ZoLA-Multi-Event}.
Another line of work leverages personalized video generation to synthesize multiple event clips with consistent subjects~\cite{VideoStudio,VideoDirectorGPT}.
To get the final video they have to concatenate all the generated clips into one, leading to abrupt scene cuts.
In addition, all existing methods present each event with a fixed-length video and cannot control the duration of individual events.

%
%
Recent work~\cite{GLIGEN,LLMGroundedDM} has shown that text-guided models often struggle with intricate spatial prompts, which can be improved by binding objects to spatial inputs (\eg bounding boxes).
Similarly, we hypothesize that the absence of explicit \emph{temporal binding} precludes successful multi-event video generation in current models.
Given a multi-event text prompt without timestamps, the generator must plan the time range of each event to form a video, which involves complicated reasoning.
Inspired by the content-motion decomposition paradigm in video generation~\cite{MoCoGAN,MotionContentVP}, we propose to use \textbf{(i)} a global caption depicting content such as background and subject appearances, and \textbf{(ii)} a sequence of \emph{temporal captions}~\cite{ActivityNetCaption} describing dynamic events, as our model input.
Each temporal caption consists of a text description and the \emph{start} and \emph{end time} of the event.
By providing temporally localized captions, the model can focus on one event at a time.
In addition, our model processes all text prompts to generate a video in one shot, which ensures consistent subjects and smooth transitions between events.

%
%
Our resulting method, named \textbf{Min}d the \textbf{T}ime (\algoNameFull), is a temporally-grounded video generator built upon a pre-trained latent Diffusion Transformer (DiT)~\cite{DiT}.
In each DiT block, we employ two cross-attention layers for global and temporal captions, respectively.
To condition the model on a sequence of events, we concatenate the text embeddings of all temporal captions and perform cross-attention.
The key challenge here is how to use the event timestamps to associate each caption with corresponding video tokens.
Inspired by Rotary Position Embedding (RoPE)~\cite{RoPE}, we introduce Rescaled RoPE (ReRoPE) to guide the event caption to focus on attending to frames within its time range while ensuring a smooth transition between adjacent events.

In summary, this work makes four main contributions:
\textbf{(i)} \algoNameFull, the first video generator that supports sequential event generation with time control.
\textbf{(ii)} A novel training strategy that conditions the model on scene cuts, facilitating training on long videos and shot transition control.
\textbf{(iii)} State-of-the-art multi-event video generation results in both text-only and image-conditioned settings on a hold-out set of our dataset and StoryBench~\cite{StoryBench}.
\textbf{(iv)} An LLM-based prompt enhancer that extends short prompts to detailed global and temporal captions, from which we can generate videos with richer motion evaluated by VBench~\cite{VBench}.

\section{Related Work}
\label{sec:related-work}

\heading{Text-to-video diffusion models.}
With recent progress in diffusion models~\cite{DiffusionModels,DDPM,FlowMatching}, text-to-video generation has achieved tremendous progress~\cite{VideoDiffusionModels,SVD,ImagenVideo}.
Earlier works inflate pre-trained image diffusion models by inserting temporal attention layers~\cite{AlignYourLatents,Make-A-Video,AnimateDiff,VideoCrafter1,VideoCrafter2,ModelScope,Tune-a-video,Text2VideoZero,VideoComposer,LVDM,MagicVideo,Lumiere}.
They typically adopt a U-Net~\cite{U-Net} model as the denoising network and run the diffusion process in a compressed latent space produced by a Variational Autoencoder (VAE)~\cite{LDM,VAE,MAGVITv2}.
Recently, Transformer-based architecture~\cite{Transformer,DiT} has drawn increasing attention as it demonstrates better scalability in generating high-resolution and complex videos~\cite{Sora,Latte,SnapVideo,CogVideoX,WALT,GenTronDiTVideo,GoogleVeo,MetaMovieGen}.
Nevertheless, we identify the inability to generate sequential events as a common failure case in these models.
By binding event captions to time and fine-tuning on temporally-grounded data, \algoNameFull greatly improves multi-event synthesis.

\heading{Story visualization.}
Traditionally, the goal of story visualization is to generate a sequence of images with consistent entities given multiple text prompts~\cite{StoryVisImg1,StoryVisImg2,StoryVisImg3,StoryVisImg4,StoryVisImg5,StoryVisImg6}.
Some recent works enhance the task by generating a video for each text prompt~\cite{MovieDreamer,StoryDiffusion,Anim-Director,TaleCrafter,Animate-A-Story}.
They usually leverage LLMs to plan the temporal ordering of events, and then run video personalization methods to generate clips with consistent character identities.
However, these methods simply concatenate all generated clips to form a story, resulting in abrupt scene cuts between events~\cite{VideoDirectorGPT,VideoStudio}.
In this work, we tackle a different task that aims to generate videos of multiple events with natural transitions.

\begin{figure*}[!t]
    \vspace{\pagetopmargin}
    \vspace{-2mm}
    \centering
    \includegraphics[width=0.98\linewidth]{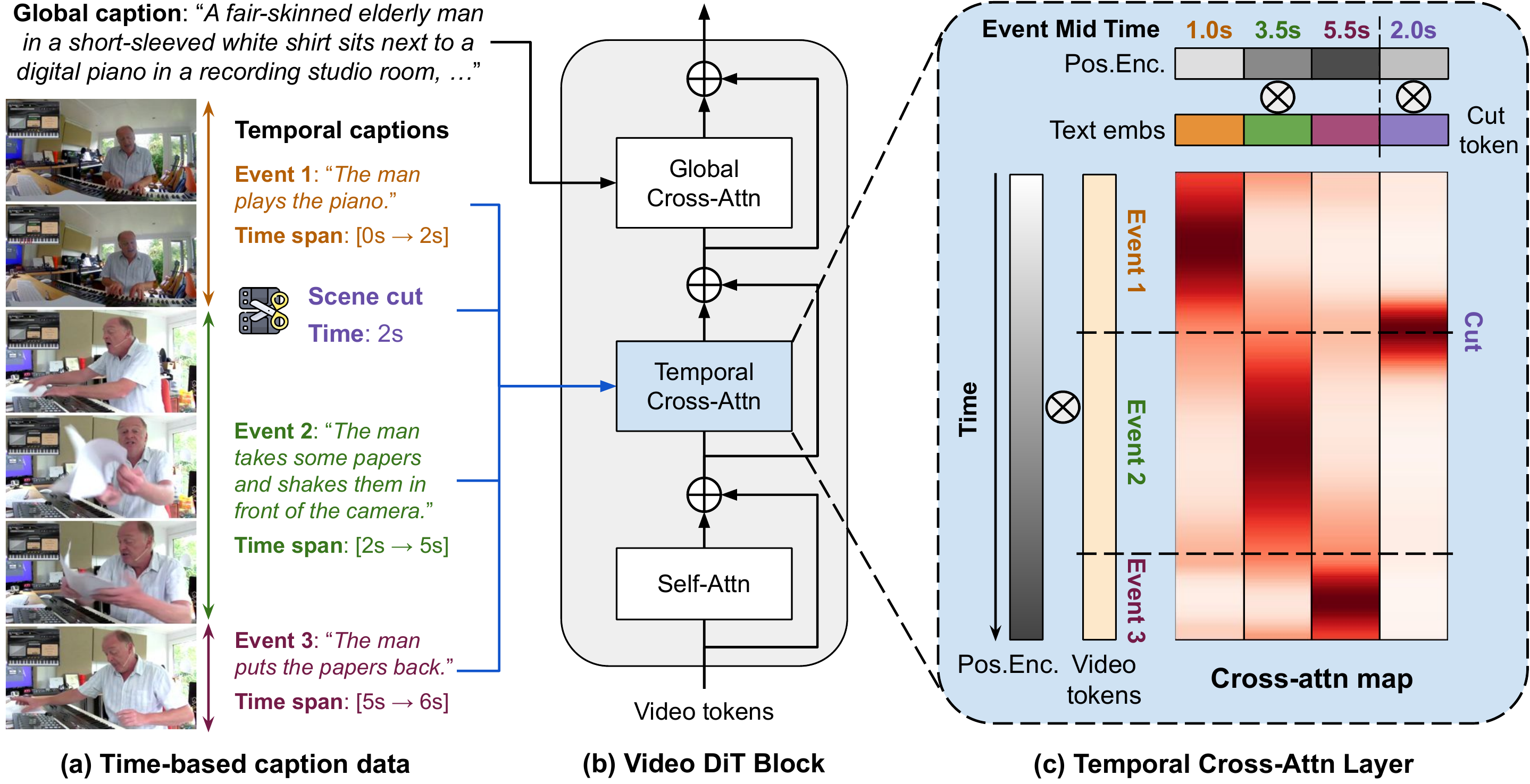}
    \vspace{\figcapmargin}
    \caption{
        \textbf{\algoNameFull framework.}
        (a) Our model takes in a global caption describing the overall video, and a list of temporal captions specifying the sequential events.
        We bind each event to a time range, enabling temporal control of the generated events.
        (b) To condition the video DiT on temporal captions, we introduce a new temporal cross-attention layer in each DiT block, which (c) concatenates the text embedding of all event prompts and leverages a time-aware positional encoding (Pos.Enc.) method to associate each event to its corresponding frames based on the event timestamps.
        \algoNameFull supports an additional scene cut conditioning, which can control the shot transition of the video.
    }
    \label{fig:pipeline}
    \vspace{\figmargin}
\end{figure*}


\heading{Multi-event video generation.}
Several studies have explored the generation of temporally consistent videos from multiple text prompts~\cite{ReuseandDiffuse,Gen-L-Video,ZoLA-Multi-Event,TALC}.
The pioneering work, Phenaki~\cite{Phenaki}, trains a masked Transformer to generate each event conditioning on its text prompt and frames from the preceding event.
However, the autoregressive generation paradigm inevitably leads to quality degradation over longer sequences.
FreeNoise~\cite{FreeNoise} and MEVG~\cite{MEVG} instead use previously generated clips to initialize the noise latent of the current clip, serving as a soft guidance for the model.
A fundamental limitation of sequential generation approaches is that they generate all events with a fixed length~\cite{StreamingT2V}.
In addition, their model lacks information about future events when generating the current event, preventing it from planning the entire video.
On the contrary, \algoNameFull processes text prompts of all events together, allowing fine-grained control of event durations and generating globally coherent videos.

\heading{Rich captions for video generation.}
Previous large-scale video-text datasets usually comprise videos with short captions~\cite{Panda-70M,WebVid}.
Recent studies have shown that detailed captions are crucial for high-quality video generation~\cite{MiraData,CogVideoX,Sora,ShareGPT4Video}.
Yet, these datasets mainly focus on the appearance and spatial layout of all entities in a video.
Closer to our task is the LVD-2M dataset~\cite{LVD-2M-Temporal-Dense-Caption-Dataset}, which labels sequential events in motion-rich videos.
However, they only use text to describe the order of events, without localizing them in time.
In this work, we are the first to enhance captions with precise timestamps for video generation.
In addition, we study a previously overlooked scene cut annotation of video data, which further enhances the controllability of our model.

\section{Method}
\label{sec:method}

\heading{Task formulation.}
Given a sequence of $\numevents$ temporally localized text prompts, $\{(\densecaption_n, \startt_n, \endt_n)\}_{n=1}^{\numevents}$, and $\numcuts$ cut timestamps, $\{\cutscenet_n\}_{n=1}^{\numcuts}$, our goal is to generate a video containing all events following their text prompt $\densecaption_n$ at the desired time range $(\startt_n, \endt_n)$.
The video is assumed to have no shot transition except at the input cut timestamps.

\heading{Overview.}
We build upon a pre-trained text-to-video Diffusion Transformer (DiT)~\cite{DiT} (\cref{sec:background}).
Our method, \algoNameFull, incorporates a temporally-aware cross-attention layer to enable event timestamp control (\cref{sec:temporal-cross-attn}) and conditioning on video scene cuts (\cref{sec:cutscene-condition}).
Finally, we design a prompt enhancer that allows users to generate multi-event videos from simple prompts with our model (\cref{sec:prompt-enhancer}).

\subsection{Background: Text-to-Video Latent DiT}
\label{sec:background}

Given a video, our latent DiT~\cite{DiT} first encodes it to video tokens $\videotokens$ with a tokenizer~\cite{VAE}.
Then, it adds Gaussian noise $\noise_t$ to $\videotokens$ to obtain a noisy sample $\videotokens_t$, and trains a denoising network following the rectified flow formulation~\cite{FlowMatching,RFSampler}:
\begin{equation}
\label{eq:dit-loss}
    \mathcal{L}_{\mathrm{DiT}} = ||\velocity_t - \ditmodel(\videotokens_t, t, \generallabel)||^2,\ \mathrm{where\ \ } \velocity_t = \noise_t - \videotokens.
\end{equation}
Here, $\ditmodel$ is implemented as a Transformer model~\cite{Transformer} consisting of a stack of DiT blocks, and $\generallabel$ denotes the conditioning signals such as text embeddings of a video caption.
Similar to recent works~\cite{WALT,MetaMovieGen,OpenSora}, each DiT block in our base model contains a self-attention layer over video tokens, a cross-attention layer fusing video and text, and an MLP.

\heading{Rotary Position Embedding (RoPE).}
To indicate the position of video tokens in attention, our base model utilizes RoPE~\cite{RoPE} due to its wide application in recent works~\cite{FiT-RoPE,LuminaT2X,CogVideoX,OpenSoraPlan}.
At a high level, given a sequence of $N$ vectors $\{\bm{x}_n\}_{n=1}^N$, RoPE computes an angle $\ropeangle_n$ for each vector $\bm{x}_n$ using its position $n$, and rotates $\bm{x}_n$ with $\ropeangle_n$ to obtain $\tilde{\bm{x}}_n$:
\begin{equation}
\label{eq:basic-rope}
    \ropeangle_n = n \ropebase,\ \ 
    \tilde{\bm{x}}_n = \mathrm{RoPE}(\bm{x}_n, n) = \bm{x}_n e^{i \ropeangle_n},
\end{equation}
where $\ropebase$ is a pre-defined base angle.\footnote{In fact, RoPE uses a list of angles $\bm{\ropeangle} \in \mathbb{R}^{h/2}$ to rotate each element of a vector $\bm{x} \in \mathbb{R}^h$ separately. We treat it as a single angle for simplicity in this paper, as all dimensions of $\bm{\ropeangle}$ changes monotonically with $n$~\cite{RoPE}.}
With RoPE, a vector $\bm{x}_n$ has a similar rotation angle with a vector $\bm{x}_m$ when $n$ and $m$ are close.
Consequently, RoPE encourages nearby vectors to have a higher self-attention weight $A_{n,m}$:
\begin{align}
\label{eq:rope-diff-angle}
\hspace{-2mm}
    A_{n, m} &= \mathrm{Re}[\dotproduct{\tilde{\bm{x}}_n}{\tilde{\bm{x}}_m}] = \mathrm{Re}[\dotproduct{\bm{x}_n}{\bm{x}_m} e^{i(n - m)\ropebase}] \notag \\
    &= \mathrm{Re}[\mathrm{RoPE}(\dotproduct{\bm{x}_n}{\bm{x}_m}, n - m)],
\end{align}
where $A_{n, m}$ decreases monotonically with $\absvalue{n - m}$ when $(n - m)\ropebase \in [-\pi/2, \pi/2]$.
This usually holds in our DiT since the video tokens $\videotokens$ are of low resolution.
Please refer to Appendix~\appref{app:rerope-property-proof} for a rigorous discussion.
In our video DiT, RoPE is only applied in the self-attention.
There is no positional encoding in the video-text cross-attention as the input text prompt is expected to describe the entire video.

\subsection{Temporally Aware Video DiT}
\label{sec:temporal-cross-attn}

Existing text-guided video diffusion models only take in one global text prompt for a video.
As shown in \cref{fig:pipeline} (a), we further input a sequence of temporal captions that bind each event to an exact time range.
The decomposition of global and temporal captions resembles the classic content-motion disentanglement in video generation~\cite{MoCoGAN,MotionContentVP}, providing a clearer guidance of video dynamics to the model.

\heading{Temporal cross-attention.}
To condition \algoNameFull on temporal captions, we add a new temporal cross-attention layer between the original self-attention and cross-attention layers as shown in \cref{fig:pipeline} (b).
Prior works~\cite{GLIGEN,Boximator,InstanceDiffusion} show that such design enables fast adaptation to new spatial conditioning input, and we show that it also works for temporal conditioning.
We first extract text embeddings $\densecaptionembs_n \in \mathbb{R}^{L^\densecaption \times D^\densecaption}$ for each event text prompt $\densecaption_n$, where $L^\densecaption$ and $D^\densecaption$ are the text length and the embedding dimension, respectively.
Then, we apply positional encoding to each $\densecaptionembs_n$ to indicate its time span $[\startt_n, \endt_n]$, and concatenate them along the sequence dimension to perform cross-attention with video tokens:
\begin{align}
\label{eq:temporal-cross-attn}
    \ropetextembs^{\densecaption}_n &= \mathrm{Pos.Enc.}(\densecaptionembs_n, \startt_n, \endt_n), \notag \\
    \tilde{\videotokens} &= \mathrm{XAttn}(\videotokens, \mathrm{Concat}([\ropetextembs^{\densecaption}_1, \ropetextembs^{\densecaption}_2, ..., \ropetextembs^{\densecaption}_{\numevents}])).
\end{align}
Apart from positional encoding, an intuitive way to indicate event time range is hard masking, where we only allow $\densecaptionembs_n$ to attend to video tokens within $[\startt_n, \endt_n]$.
However, for frames close to an event transition point, it is beneficial to receive information from both events to synthesize a smooth transition.
Therefore, we decide to use RoPE to serve as soft masking to guide the text embedding of each event.

Intuitively, we want the temporal cross-attention to have three key properties:
\textbf{(i)} For video tokens within the time span of an event, they should always attend the most to the text embedding of this event.
\textbf{(ii)} For an event, the attention weight should peak with the video token at the midpoint of its time span, and then decrease towards the boundary of the event.
\textbf{(iii)} The video token at the transition point between two events should attend equally to their text embeddings, which helps the model localize the event boundary.
\\
Below, we show that vanilla RoPE fails to achieve (i) and (iii), necessitating a new positional encoding for this task.

\begin{figure}[!t]
    \vspace{\pagetopmargin}
    \centering
    \hspace{-3mm}\includegraphics[width=1.03\linewidth]{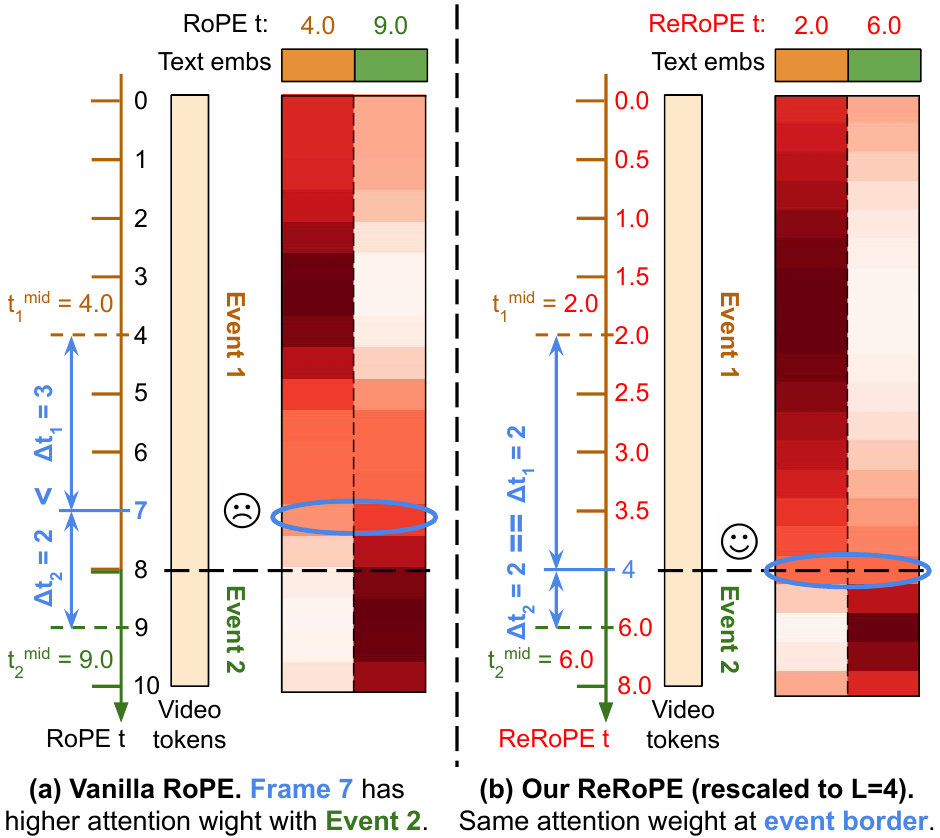}
    \vspace{\figcapmargin}
    \vspace{-4mm}
    \caption{
        \textbf{Comparison of vanilla RoPE and our Rescaled RoPE.}
        We use the same random vector for video tokens and text embeddings to only visualize the bias introduced by positional encoding.
        (a) Vanilla RoPE uses raw timestamps as the rotation angle, where frames within one event might be biased to the wrong text.
        (b) We instead rescale all events to have the same length $L$, so that video tokens always attend the most to the current event.
        In addition, frames at event boundaries attend to adjacent events equally.
    }
    \label{fig:rerope-vis}
    \vspace{\figmargin}
    \vspace{1mm}
\end{figure}


\heading{Vanilla temporal RoPE.}
We start from the standard RoPE in \cref{eq:basic-rope}.
For a video token $\videotokens_{[t, \cdot, \cdot]}$ at any spatial location on frame $t$, we only use the timestamp $t$ to determine its rotation angle $\ropeangle$ since we focus on temporal correspondence here.
For an event happening in $[\startt_n, \endt_n]$, a natural way to encode its text embedding is using its middle timestamp $\midt_n = (\startt_n + \endt_n) / 2$.
Therefore, the vanilla RoPE is as:
\begin{align}
\label{eq:vanilla-rope}
    \tilde{\videotokens}_{[t, \cdot, \cdot]} = \mathrm{RoPE}(\videotokens_{[t, \cdot, \cdot]}, t),\ \ 
    \ropetextembs^{\densecaption}_n = \mathrm{RoPE}(\textembs^{\densecaption}_n, \midt_n), \\
    \mathrm{Attn}(\tilde{\videotokens}_{[t, \cdot, \cdot]}, \ropetextembs^{\densecaption}_n) = \mathrm{Re}[\mathrm{RoPE}(\dotproduct{\videotokens_{[t, \cdot, \cdot]}}{\textembs^{\densecaption}_n}, t - \midt_n)]
\label{eq:vanilla-rope-angle-diff}
\end{align}
Such design satisfies property (ii), while violating the other properties as shown in \cref{fig:rerope-vis} (a).
In this example, frame 7 belonging to the first event is closer to $\midt_2$ than $\midt_1$ and thus has a higher attention weight with the second event.
In addition, frame 8 which is at the intersection of two events attends to the second event more than the first one.
As a result, the model cannot locate the correct event boundary.

\heading{Rescaled RoPE (ReRoPE).}
When adjacent events have different durations, their midpoints' distance to the event boundary also becomes different, causing vanilla RoPE to fail.
Therefore, we propose to rescale all events to the same length $L$ and recompute timestamps for encoding in \cref{eq:vanilla-rope}.
For a timestamp $t$ lying in the $n$-th event, we transform it as:
\begin{equation}
\label{eq:rerope-rescaled-t}
    \tilde{t} = \frac{(t - \startt_n) L}{\endt_n - \startt_n} + (n - 1) L,\ \ \mathrm{s.t.}\ \startt_n \leq t \leq \endt_n.
\end{equation}
Using \cref{eq:rerope-rescaled-t} for both video tokens and events, we have:
\begin{equation}
\label{eq:rerope-angle-diff}
    \tilde{t} - \tilde{t}^{\mathrm{mid}}_n = \left (\frac{t - \startt_n}{\endt_n - \startt_n} - \frac{1}{2} \right ) L.
\end{equation}
As we show in Appendix~\appref{app:rerope-property-proof}, our ReRoPE design achieves all three desired properties in the temporal cross-attention.

Inspired by Positional Interpolation~\cite{PI-RoPE}, we set a fixed value for $L$, so that videos of different lengths are rescaled to the same length in \cref{eq:rerope-angle-diff}.
As a result, ReRoPE always induces the same attention bias to temporal cross-attention, making the layer invariant to the actual video length.

\subsection{Scene Cut Conditioning}
\label{sec:cutscene-condition}

Prior large-scale video datasets usually remove videos with scene cuts or split them into shorter clips~\cite{Panda-70M,MiraData,InternVid}.
Indeed, training a generator on videos with cuts may lead to undesired scene transitions in generated videos.

Typically, professionally edited videos contain frequent cuts, and excluding them in training may lose valuable information.
Removing such clips also reduces the amount of training data significantly (in our data 20\% of clips contain cuts).
But most importantly, it makes a model unable to use such a valuable cinematographic effect, leading to temporally cropped videos.
Prior image generators face a similar issue with image cropping~\cite{SDXL}, where the model may learn to generate ``cropped" images with out-of-frame objects.
Based on these insights, we decide to keep all the videos, while explicitly conditioning the model on the timestamps of cuts.
Once the model learns such conditioning, we can input zeros during inference to enforce a cut-free video.

We treat a scene cut as a special event with the same content and equal start and end timestamps.
To condition \algoNameFull on it, we initialize a learnable vector $\cuttoken \in \mathbb{R}^{1 \times D^\densecaption}$, apply ReRoPE with its timestamp $\cutscenet_n$ transformed by \cref{eq:rerope-rescaled-t}, and concatenate it with the text embeddings of temporal captions to perform cross-attention with video tokens:
\begin{align}
\label{eq:temporal-cross-attn-w-cuts}
    \ropecuttoken_n &= \mathrm{ReRoPE}(\cuttoken_n, \cutscenet_n, \cutscenet_n), \notag \\
    \hspace{-2mm}
    \tilde{\videotokens} = \mathrm{XAttn}(\videotokens,\ &\mathrm{Concat}([\ropetextembs^{\densecaption}_1, ..., \ropetextembs^{\densecaption}_{\numevents}, \ropecuttoken_1, ..., \ropecuttoken_{\numcuts}])).
\end{align}
As we show in the ablation (\cref{sec:ablation}), this design greatly reduces undesired scene transitions when they are not requested, and allows a practitioner to use them when needed.

\subsection{Prompt Enhancer}
\label{sec:prompt-enhancer}

\algoNameFull offers video generation with precise control of event timing. Yet, in certain applications starting from a single prompt can be more desirable. Prior works demonstrated that LLMs can generate physically meaningful spatial layout of scenes from text prompts~\cite{LLMGroundedDM,VideoDirectorGPT}. Similarly, we show that LLMs can plan the temporal structure of multi-event videos. 
Given a short text, we prompt LLMs to extend it to a detailed global caption and several event captions with their time span.
Then, our model can generate a video with rich motion content from the enhanced prompts.

\section{Experiments}
\label{sec:experiments}

Our experiments aim to answer the following questions:
\textbf{(i)} Can \algoNameFull control event timing in both text-to-video (T2V) and image-to-video (I2V) settings? (\cref{sec:t2v-results} \& \cref{sec:i2v-results})
\textbf{(ii)} Does prompt enhancement lead to high-quality multi-event videos from a single prompt? (\cref{sec:llm-enhanced-results})
\textbf{(iii)} What is the impact of each design choice in our framework? (\cref{sec:ablation})

\begin{figure*}[!t]
    \vspace{\pagetopmargin}
    \vspace{-2mm}
    \centering
    \includegraphics[width=\linewidth]{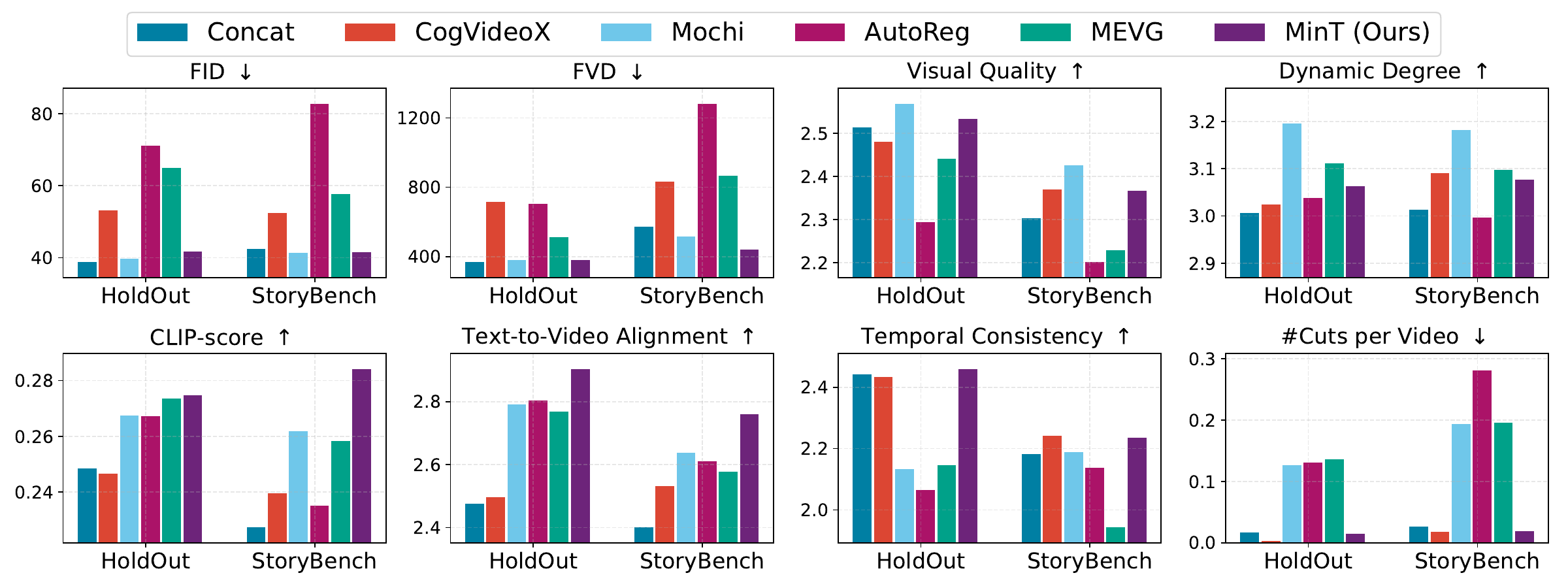}
    \vspace{\figcapmargin}
    \vspace{-6.5mm}
    \caption{
        \textbf{T2V results on \holdout and StoryBench.}
        For CogVideoX and Mochi we concatenated the events into a single prompt, similar to the \jointmodel baseline. Metrics in the first row measure visual quality, while those in the second row focus on the text alignment and transition smoothness between events.
        \algoNameFull performs the best in event-related metrics while maintaining a high visual quality.
    }
    \label{fig:t2v-quan-results}
    \vspace{\figmargin}
\end{figure*}


\subsection{Experimental Setup}
\label{sec:exp-setup}

We list some key aspects of our experimental setup here.
For full details, please refer to Appendix~\appref{app:more-exp-setup}.

\heading{Training data.}
Existing video datasets with time-based captions usually come from dense video captioning~\cite{ActivityNetCaption,YouCook2}.
However, these datasets are limited in scale, which are impossible to fine-tune a large-scale video generator on.
Therefore, we manually annotate temporal events on videos sourced from existing datasets~\cite{Panda-70M,HD-VILA}, resulting in around 200k videos, where we hold out 2k videos for evaluation.
To condition the model on scene cuts, we run TransNetV2~\cite{TransNetV2} to detect scene boundaries on annotated videos.

\heading{Evaluation datasets.}
We leverage the 2k holdout videos as our primary benchmark (dubbed \textit{\holdout}).
We also test on the \textit{StoryBench}~\cite{StoryBench} dataset, which annotates temporal captions similar to ours.
We filter out videos with only a single event, leading to around 3k testing samples.
Finally, to test \algoNameFull's ability in generating motion-rich videos from short prompts, we utilize prompt lists from \textit{VBench}~\cite{VBench}.

\begin{figure*}[!t]
    \vspace{\pagetopmargin}
    \vspace{-2mm}
    \centering
    \includegraphics[width=\linewidth]{imgs/qual/t2v/t2v-qual-ex1.pdf}
    \vspace{\figcapmargin}
    \vspace{-4.5mm}
    \caption{
        \textbf{Qualitative results of T2V.}
        Concatenating all events into a single prompt (\jointmodel) can only generate the first event.
        Autoregressive generation (\imgcondmodel) suffers from video stagnation and cannot generate the third event.
        MEVG struggles to preserve the person's identity and produces abrupt event transitions.
        \algoNameFull is the only method that generates all events with smooth transitions and consistent content.
        See Appendix~\appref{app:more-qual-comparison} for more qualitative results and our \href{https://mint-video.github.io/\#more-our-results}{project page} for video results.
    }
    \label{fig:t2v-qual-results}
    \vspace{\figmargin}
\end{figure*}

\heading{Baselines.}
To show that current video models are not capable of generating multi-event videos, we design a straightforward method, called \textit{\jointmodel}, that simply concatenates all the prompts together.
We apply it to both our base model and state-of-the-art open-source models CogVideoX~\cite{CogVideoX} and Mochi~\cite{Mochi}.
We also compare to approaches with code available and are designed to generate smoothly connected events.
\textit{MEVG}~\cite{MEVG} is the state-of-the-art multi-event video generation method.
It generates each event from its prompt separately.
To ensure smooth transitions, it runs DDIM inversion~\cite{DDIM} on the previously generated event as the noise initialization for the current event.
We also design a baseline that fine-tunes an image-conditioned video diffusion model to generate events autoregressively (dubbed \textit{\imgcondmodel}).
To separate the impact of architecture from the method, we implement both MEVG and \imgcondmodel on top of our base model ensuring a fair comparison.
Notably, since no baselines can control event timing, we simply set all events to have the same length to make the comparison possible.

\heading{Evaluation metrics.}
We focus on three dimensions: visual quality, text alignment, and event transition smoothness.
We report common metrics such as FID~\cite{FID}, FVD~\cite{FVD} for visual quality, and per-frame CLIP-score~\cite{CLIPScore} for text alignment.
In addition, we leverage a state-of-the-art video quality assessment model, VideoScore~\cite{VideoScore} as it has been shown to produce results consistent with human evaluators.
We take the \textit{visual quality} and \textit{dynamic degree} output for visual quality, the \textit{text-to-video alignment} output for text alignment, and the \textit{temporal consistency} output for event transition smoothness.
Notably, since we care about event generation, we compute text alignment between temporal captions and video clips cropped out based on the event span.
Finally, we run TransNetV2 to detect the cuts in generated videos to measure event transition smoothness.

\heading{Implementation details.}
\algoNameFull builds upon a pre-trained latent video DiT similar to~\cite{OpenSoraPlan,OpenSora}.
It generates videos of 512$\times$288 resolution and up to 12 seconds.
We fine-tune the entire model with the AdamW optimizer~\cite{AdamW} and a batch size of 512 for 12k steps.
For inference, we run 256 denoising steps with a classifier-free guidance~\cite{CFG} scale of 8.

\subsection{Text-to-Video Generation}
\label{sec:t2v-results}

\cref{fig:t2v-quan-results} presents the quantitative results on \holdout and StoryBench datasets.
\cref{fig:t2v-qual-results} shows a qualitative comparison.
Compared to \jointmodel which shares the same base model as ours, \algoNameFull achieves slightly lower visual quality on \holdout and better results on StoryBench.
This is because StoryBench prompts are out-of-distribution to our model.
Despite this, time-based captions help \algoNameFull generate a video with a good temporal structure.
On the other hand, we generate events with much higher text alignment.
We draw similar observations when comparing \algoNameFull to the \jointmodel baseline based on CogVideoX and Mochi.
Overall, this proves that our model acquires the new capability of sequential event generation while maintaining high visual quality.

As for multi-event generation methods, \imgcondmodel and MEVG greatly improve the text alignment, as they generate each event from its prompt separately.
Yet, \imgcondmodel has much lower visual quality, since conditioning on generated frames leads to artifacts such as video stagnation.
MEVG resolves this issue with frame inversion.
However, it often generates abrupt transitions between events as indicated by the large number of cuts.
In fact, we found that the inversion technique in MEVG only works well when two consecutive event captions have a similar structure, \eg the same subject doing different actions.
When two captions have a subject change such as in \cref{fig:t2v-qual-results}, the generated events usually contain completely different characters.
Overall, \algoNameFull achieves the best balance between video quality, event localization, and temporal smoothness of the video.
See Appendix~\appref{app:more-compare-with-sota} for comparisons with commercial models.

\begin{figure}[!t]
    \centering
    \includegraphics[width=\linewidth]{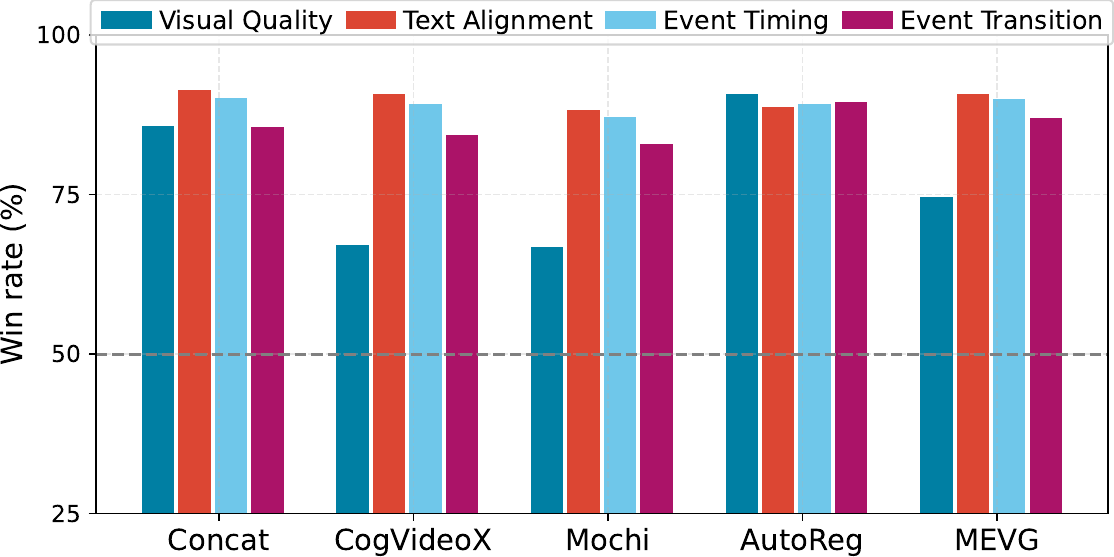}
    \vspace{\figcapmargin}
    \vspace{-4mm}
    \caption{
        \textbf{Human preference evaluation against T2V baselines.}
        \algoNameFull is better or competitive in visual quality, and significantly outperforms baselines in the other three event-related metrics.
    }
    \label{fig:t2v-user-study}
    \vspace{\figmargin}
\end{figure}

\heading{Human evaluation.}
We conduct a user study using 200 randomly sampled prompts from \holdout.
We ask the participants to express their preference when presented with paired samples from \algoNameFull and each baseline, gathering votes from 20 users per sample.
Results in \cref{fig:t2v-user-study} show that \algoNameFull has better or competitive visual quality, while generating events with significantly higher text alignment, timing accuracy, and transition smoothness.

\heading{Event time control.}
\algoNameFull supports fine-grained control of event timing.
Please refer to Appendix~\appref{app:event-time-control} for our results.

\subsection{Image-conditioned Video Generation}
\label{sec:i2v-results}

We evaluate the model's ability to animate entities in an existing image to perform sequential events.
Following~\cite{StoryBench}, models have access to the ground-truth initial frame of testing videos as well as the event text prompts.

\heading{Settings.}
The same datasets and metrics as in the T2V setting are employed.
We compare with the best baseline, \textit{MEVG}.
It has an image-conditioned variant that replicates the initial frame to form a pseudo video.
For \algoNameFull, we fine-tune it to condition on an image by concatenating the image with the noisy latent similar to prior works~\cite{SVD,DynamiCrafter}.

\heading{Results.}
\cref{tab:i2v-quan-results} presents the multi-event image animation results on \holdout and StoryBench datasets.
We draw similar observations as in the T2V setting.
\algoNameFull achieves either better or competitive results in visual quality, while performing significantly better in text alignment with event captions and temporal smoothness of event transitions.

\begin{table}[t]
    \vspace{1mm}
    \centering
    \setlength{\tabcolsep}{3.5pt}
    \scriptsize
    \begin{tabular}{l|cccc|cc|cc}
        \toprule
        \textbf{Method} & FID $\downarrow$ & FVD $\downarrow$ & VQ $\uparrow$ & DD $\uparrow$ & CLIP-T $\uparrow$ & TA $\uparrow$ & TC $\uparrow$ & \#Cuts $\downarrow$ \\
        \midrule
        \multicolumn{9}{c}{\textit{Dataset: \holdout}} \\
        \midrule
        MEVG & 57.57 & 495.75 & 2.56 & \textbf{3.39} & 0.266 & 2.72 & 2.25 & 0.108 \\
        \textbf{Ours} & \textbf{22.04} & \textbf{218.21} & \textbf{2.60} & 3.30 & \textbf{0.272} & \textbf{3.00} & \textbf{2.47} & \textbf{0.025} \\
        \midrule
        \multicolumn{9}{c}{\textit{Dataset: StoryBench}} \\
        \midrule
        MEVG &  56.51 & 732.94 & 3.27 & \textbf{3.80} & 0.265 & 2.83 & 3.03 & 0.150 \\
        \textbf{Ours} & \textbf{21.85} & \textbf{314.59} & \textbf{3.36} & 3.76 & \textbf{0.273} & \textbf{3.37} & \textbf{3.29} & \textbf{0.014} \\
        \bottomrule
    \end{tabular}
    \vspace{\tablecapmargin}
    \caption{
        \textbf{I2V results on \holdout and StoryBench.}
        VQ, DD, TA, and TC stand for visual quality, dynamic degree, text-to-video alignment, and temporal consistency from VideoScore.
        \#Cuts is the average number of cuts per video.
        Similar to T2V, \algoNameFull also achieves better visual quality and smooth event transition.
    }
    \label{tab:i2v-quan-results}
    \vspace{\tablemargin}
\end{table}


\subsection{Prompt Enhanced Video Generation}
\label{sec:llm-enhanced-results}

\algoNameFull introduces a new dimension to prompt enhancement, where users can control the amount of motion in the generated video via temporal captions.
We show that this process can be automated by an LLM.
This enables users to generate more interesting videos from a short prompt.

\heading{Dataset.}
Since we are interested in the motion of generated videos, we take the list of prompts from the \textit{Dynamic Degree} evaluation dimension on VBench~\cite{VBench}.
These prompts are diverse and always contain subjects performing non-static actions.
Yet, they are all short with around 10 words.

\begin{table}[t]
    \vspace{\pagetopmargin}
    \vspace{-1mm}
    \centering
    \setlength{\tabcolsep}{2.5pt}
    \scriptsize
    \begin{tabular}{l|cccc|c|c}
        \toprule
        \multirow{2}{*}{\textbf{Method}} & Subject & Background & Aesthetic & Imaging & Motion & Dynamic \\
        & Consist. $\uparrow$ & Consist. $\uparrow$ & Quality $\uparrow$ & Quality $\uparrow$ & Smooth $\uparrow$ & Degree $\uparrow$ \\
        \midrule
        Short &  0.857 & 0.939 & 0.498 & 0.583 & 0.995 & 0.481 \\
        Global & 0.890 & 0.950 & 0.541 & 0.613 & 0.995 & 0.517 \\
        \midrule
        \textbf{Ours} & 0.900 & 0.950 & 0.544 & 0.609 & 0.988 & \textbf{0.711} \\
        \bottomrule
    \end{tabular}
    \vspace{\tablecapmargin}
    \caption{
        \textbf{Prompt enhancement results on VBench.}
        Consist. means consistency.
        The first four metrics measure video quality, while we focus on the motion of generated videos.
        \algoNameFull generates videos with significantly higher dynamics degree and competitive visual quality and motion smoothness.
    }
    \label{tab:vbench-enhancer-quan-results}
    \vspace{\tablemargin}
\end{table}


\heading{Prompt enhancer.}
We prompt GPT-4~\cite{GPT-4} to extend the short prompt to a detailed global caption and temporal captions.
Please refer to Appendix~\appref{app:prompt-enhancement-vbench} for the prompt we use.

\heading{Baselines and evaluation metrics.}
We compare with videos generated by our base model using the original short prompts (dubbed \textit{Short}).
To disentangle the effect of global caption and temporal captions, we also compare to videos generated by our base model using the enhanced global caption (named \textit{Global}).
For evaluation, we compute six metrics from the official VBench test suite, which focus on visual quality, temporal smoothness, and motion richness.

\begin{figure}[!t]
    \vspace{1mm}
    \centering
    \includegraphics[width=1.0\linewidth]{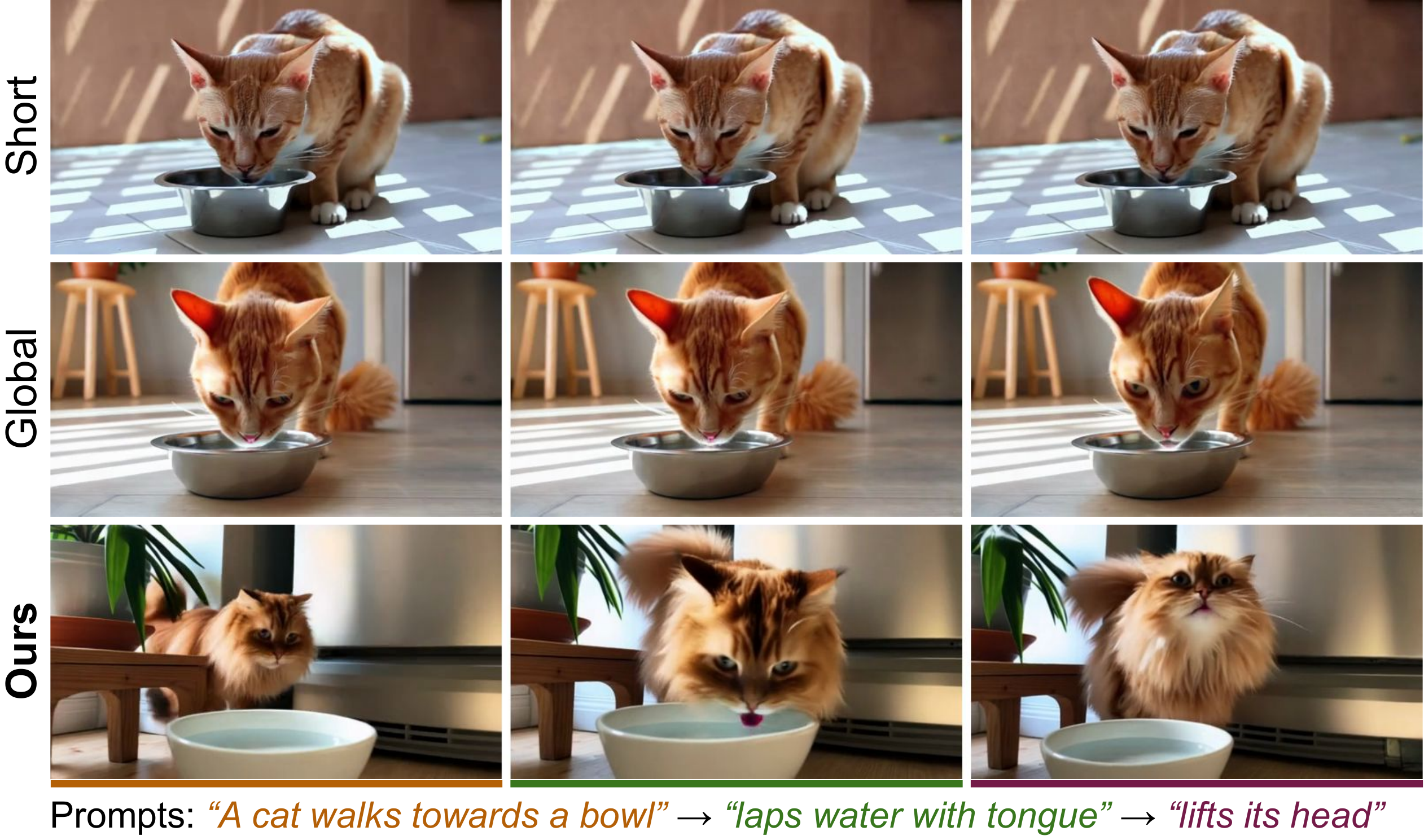}
    \vspace{\figcapmargin}
    \vspace{-4mm}
    \caption{
        \textbf{Qualitative comparison of prompt enhancement results.}
        The original short prompt is ``a cat drinking water".
    }
    \label{fig:vbench-prompt-enhance-ex1}
    \vspace{\figmargin}
    \vspace{1mm}
\end{figure}

\heading{Results.}
\cref{tab:vbench-enhancer-quan-results} demonstrates the video generation results from enhanced prompts on VBench.
Global performs consistently better than Short, proving that using a detailed prompt is indeed beneficial.
When equipped with additional temporal captions, \algoNameFull achieves competitive results with baselines in visual quality and motion smoothness, while scoring significantly higher in dynamic degree.
\cref{fig:vbench-prompt-enhance-ex1} shows a qualitative example, where we turn a single-action prompt into a coherent video with three actions.
Please refer to Appendix~\appref{app:prompt-enhancement-vbench} for more qualitative results.

\subsection{Ablation Study}
\label{sec:ablation}

We study the effect of each component in our model in \cref{tab:ablation-quan-results}.
All ablations are conducted on \holdout.

\heading{Time conditioning.}
We examine different ways to condition the model on the event time span.
\textit{Concat time} runs an MLP to embed timestamps to high dimensional features, and then concatenates them with text embeddings of temporal captions.
However, since our base model uses RoPE, the video tokens do not contain absolute positional information.
Therefore, doing cross-attention with time-embedded text features cannot associate events to video frames, leading to significantly worse text alignment with event captions.
\textit{Hard attn mask} adopts hard masking in the temporal cross-attention, where events only attend to frames within its time range.
This enables synthesizing events at desired time periods.
However, hard masking prevents video tokens at event boundaries from attending to upcoming events, resulting in abrupt event transitions thus lower temporal consistency and more scene cuts.
Finally, \textit{Vanilla RoPE} encodes video tokens and text embeddings of events with raw timestamps.
As discussed in \cref{fig:rerope-vis}, it fails to accurately locate event borders, which degrades the control of event timing as indicated by the lower text alignment scores.

\heading{ReRoPE rescaling length $L$.}
By default, we set $L = 8$.
\cref{tab:ablation-quan-results} shows that using $L=4$ or $16$ achieves similar results.
This indicates that the model is insensitive to this hyper-parameter.
Please refer to Appendix~\appref{app:rerope-vis-L} for more discussions on ReRoPE with different values of $L$.

\heading{Scene cut conditioning.}
In the last row of \cref{tab:ablation-quan-results}, we remove the scene cut conditioning during training.
As discussed in \cref{sec:cutscene-condition}, without access to scene cut information, the model will introduce undesired shot transitions in the generated video.
Indeed, this variant has similar visual quality and text alignment as our full model, but scores much lower in temporal consistency and generates more cuts.
Please refer to Appendix~\appref{app:scene-cut-conditioning} for more analysis.

\begin{table}[t]
    \vspace{\pagetopmargin}
    \vspace{-1mm}
    \centering
    \setlength{\tabcolsep}{5pt}
    \scriptsize
    \begin{tabular}{l|cc|cc|cc}
        \toprule
        \textbf{Method} & VQ $\uparrow$ & DD $\uparrow$ & CLIP-T $\uparrow$ & TA $\uparrow$ & TC $\uparrow$ & \#Cuts $\downarrow$ \\
        \midrule
        \textbf{Full Model} & 2.56 & 3.32 & \textbf{0.270} & \textbf{2.92} & 2.44 & 0.026 \\
        \midrule
        Concat time & 2.53 & 3.31 & 0.249 & 2.42 & 2.33 & 0.075 \\
        Hard attn mask & 2.45 & 3.34 & 0.260 & 2.68 & 2.30 & 0.069 \\
        Vanilla RoPE & 2.54 & 3.32 & 0.262 & 2.79 & 2.42 & 0.030 \\
        \midrule
        ReRoPE ($L$=4) & 2.54 & 3.33 & 0.264 & 2.88 & 2.43 & 0.029 \\
        ReRoPE ($L$=16) & 2.55 & 3.32 & 0.265 & 2.90 & 2.44 & 0.025 \\
        \midrule
        No cut condition & 2.54 & 3.33 & 0.268 & 2.89 & 2.34 & 0.084 \\
        \bottomrule
    \end{tabular}
    \vspace{\tablecapmargin}
    \caption{
        \textbf{Ablation results on \holdout.}
        We study different conditioning mechanisms for event time span, the rescale length $L$ in ReRoPE, and the use of scene cut conditioning.
        VQ, DD, TA, and TC stand for visual quality, dynamic degree, text-to-video alignment, and temporal consistency from VideoScore.
        \#Cuts is the average number of scene cuts per video.
    }
    \label{tab:ablation-quan-results}
    \vspace{\tablemargin}
\end{table}


\section{Conclusion}
\label{sec:conclusion}

We present \algoNameFull, a framework for multi-event video generation with event timing control.
Our method employs a unique positional encoding method to guide the temporal dynamics of the video, resulting in smoothly connected events and consistent subjects.
Equipped with LLMs, we further design a prompt enhancer that can generate motion-rich videos from a simple prompt.
We view our work as an important step towards controllable content creation tools.
We discuss our limitations and failure cases in Appendix~\appref{app:failure-case}.

\subsection*{Acknowledgments}

We would like to thank Tsai-Shien Chen, Alper Canberk, Yuanhao Ban, Sherwin Bahmani, Moayed Haji Ali, and Xijie Huang for valuable discussions and support.

{
    \small
    \bibliographystyle{ieeenat_fullname}
    \bibliography{references}
}

\clearpage

\appendix

\setcounter{page}{1}
\setcounter{equation}{0}
\maketitlesupplementary

We highly encourage the readers to check out our \href{https://mint-video.github.io/}{project page} for video results of baselines and \algoNameFull.

\section{Details on Rotary Position Embedding}
\label{app:more-rope-details}

\subsection{Derivation of RoPE}

We detail the derivation conducted in \cref{sec:background} of the main paper.
Our derivation mostly follows \cite{RoPE,RoPE-Extrapolation-Scaling,YARN-RoPE} and only provides an intuitive motivation for our method.
We refer readers to their papers for more rigorous results.

Given a query vector $\bm{q}_n = \left[q_n^{(0)}, \cdots, q_n^{(d-1)}\right] \in \mathbb{R}^d$ at index $n$ and a key vector $\bm{k}_m = \left[k_m^{(0)}, \cdots, k_m^{(d-1)}\right] \in \mathbb{R}^d$ at index $m$, to apply RoPE, it first groups every two elements in them, and make them complex numbers as:
\begin{equation}
\label{app-eq:group-vector-to-complex}
\begin{gathered}
\hspace{-5mm}
    \bar{\bm{q}}_n = \left[\bar{q}_n^{(0)}, \cdots, \bar{q}_n^{(d/2-1)}\right],\ \ 
    \bar{q}_n^{(l)}=q_n^{(2l)}+iq_n^{(2l+1)}, \\
\hspace{-3mm}
    \bar{\bm{k}}_m = \left[\bar{k}_m^{(0)}, \cdots, \bar{k}_m^{(d/2-1)}\right],\ \ 
    \bar{k}_m^{(l)}=k_m^{(2l)}+ik_m^{(2l+1)}.
\end{gathered}
\end{equation}
Then, RoPE rotates each complex number by an angle $\theta_l$, which is achieved as element-wise multiplication:
\begin{equation}
\label{app-eq:apply-rotation-angle}
    \tilde{\bm{q}}_n = \bar{\bm{q}}_n \odot e^{i n \bm{\theta}},\ \ 
    \tilde{\bm{k}}_m = \bar{\bm{k}}_m \odot e^{i m \bm{\theta}},
\end{equation}
where $\bm{\theta}$ is determined by the position $l$ of each element in a vector.
We follow prior works~\cite{RoPE,OpenSoraPlan} to use:
\begin{equation}
\label{app-eq:rope-angle-theta}
    \bm{\theta} = \left[\theta_0, \cdots, \theta_{d/2-1}\right],\ \ 
    \theta_l = 10000^{-2l/d}.
\end{equation}
\cref{app-eq:rope-angle-theta} indicates that each $\theta_l$ is a fixed value, and thus the rotation results in \cref{app-eq:apply-rotation-angle} is only decided by the vectors' index $n$ and $m$.
This is why in the main paper, we only consider a single $\ropebase$ instead of $\bm{\theta}$ for different elements.

We can now calculate the attention between $\tilde{\bm{q}}_n$ and $\tilde{\bm{k}}_m$:
\begin{equation}
\label{app-eq:rope-attn-bias}
\begin{aligned}
    A_{n,m} = \mathrm{Re}&\left[\dotproduct{\tilde{\bm{q}}_n}{\tilde{\bm{k}}_m}\right] \\ 
    = \mathrm{Re}&\left[(\bar{\bm{q}}_n e^{i n \bm{\theta}}) \cdot (\bar{\bm{k}}_m e^{i m \bm{\theta}})^*\right] \\
    = \mathrm{Re}&\left[\sum_{l=0}^{d/2-1}{(\bar{q}_n^{(l)}e^{i n \theta_l}) (\bar{k}_m^{(l) *}e^{-i m \theta_l}})\right] \\
    = \mathrm{Re}&\left[{\sum_{l=0}^{d/2-1}\bar{q}_n^{(l)}} \bar{k}_m^{(l) *} e^{i (n - m) \theta_l}\right] \\
    = \sum_{l=0}^{d/2-1}
    &\left(q_n^{(2l)} k_m^{(2l)} + q_n^{(2l+1)} k_m^{(2l+1)}\right)\cos{((n - m)\theta_l)} + \\
    &\left(q_n^{(2l)} k_m^{(2l+1)} - q_n^{(2l+1)} k_m^{(2l)}\right)\sin{((n - m)\theta_l)}.
\end{aligned}
\end{equation}
Since we are interested in the bias introduced by RoPE in attention, we assume all queries $\bm{q}_n$ and all keys $\bm{k}_m$ are the same, so that their attention values without RoPE is the same.
Empirically, we find that query and key vectors indeed have similar values in our DiT due to the use of Layer Normalization~\cite{LayerNorm}.
Thanks to the periodic property of $\sin{(\cdot)}$ and $\cos{(\cdot)}$, from \cref{app-eq:rope-attn-bias}, we have $A_{n,m} = A_{m,n}$, \ie, the attention bias between $\bm{q}_n$ and $\bm{k}_m$ is only affected by the absolute distance between the two vectors, $\absvalue{n - m}$.

The original RoPE paper~\cite{RoPE} proves that the upper bound of $A_{n,m}$ decays monotonically with the distance $\absvalue{n - m}$ until around 40.
Since the RoPE used in the temporal cross-attention layer only encodes vectors using the temporal frame index, and our video DiT is trained on video tokens with up to around 50 frames, we roughly preserve the monotonicity of RoPE.
As we will see in \cref{app:rerope-vis-L}, while there are some fluctuations of $A_{n,m}$ in the long range, the long-term decay makes their values significantly low.

\subsection{Proof of the Property of ReRoPE}
\label{app:rerope-property-proof}

In \cref{sec:temporal-cross-attn} of the main paper, we propose to rescale all events to a fixed length $L$.
For a timestamp $t$ lying in the $n$-th event, we transform it as:
\begin{align}
\label{app-eq:rerope-rescaled-t}
    \tilde{t} = \frac{(t - \startt_n) L}{\endt_n - \startt_n} + &(n - 1) L,\ \ \mathrm{s.t.}\ \startt_n \leq t \leq \endt_n, \notag \\
    \tilde{t}^{\mathrm{mid}}_n &= L/2 + (n - 1) L.
\end{align}
After transformation, the distance between a video token in the $n$-th event and the middle timestamps of this event is:
\begin{equation}
\label{app-eq:rerope-angle-diff}
    \left\vert \tilde{t} - \tilde{t}^{\mathrm{mid}}_n \right \vert = \left\vert \frac{t - \startt_n}{\endt_n - \startt_n} - \frac{1}{2} \right \vert  L.
\end{equation}
Next, we prove that it satisfies the three desired properties of the temporal cross-attention:

\heading{(i)} \textit{For video tokens within the time span of an event, they should attend the most to the text embedding of this event.}
\\
\heading{Proof}\ \ For $\startt_n \leq t \leq \endt_n$, we have:
\begin{equation}
    -\frac{1}{2} \leq \left(\frac{t - \startt_n}{\endt_n - \startt_n} - \frac{1}{2} \right) \leq \frac{1}{2},
\end{equation}
thus, $\left \vert \tilde{t} - \tilde{t}^{\mathrm{mid}}_n \right \vert \leq L/2$.
For any $m$-th event with $m \neq n$, its distance to this video token is:
\begin{equation}
    \left \vert \tilde{t} - \tilde{t}^{\mathrm{mid}}_m \right \vert = \left \vert \left(\frac{t - \startt_n}{\endt_n - \startt_n} - \frac{1}{2} \right) + (n - m) \right \vert L.
\end{equation}
Since $|n - m| \geq 1$, we get:
\begin{equation}
    \left \vert \left(\frac{t - \startt_n}{\endt_n - \startt_n} - \frac{1}{2} \right) + (n - m) \right \vert  \geq \frac{1}{2}.
\end{equation}
Therefore, we have:
\begin{equation}
    \left \vert \tilde{t} - \tilde{t}^{\mathrm{mid}}_m \right \vert \geq L/2 \geq \left \vert \tilde{t} - \tilde{t}^{\mathrm{mid}}_n \right \vert,\ \ \forall\ m \neq n.
\end{equation}
Since RoPE attention decays monotonically with the distance, we reach the property.

\heading{(ii)} \textit{For an event, the attention weight should peak with the video token at the midpoint of its time span, and then decrease towards the boundary of the event.}
\\
\heading{Proof}\ \ When a video token is at the midpoint of an event, we have $\tilde{t} - \tilde{t}^{\mathrm{mid}}_n = 0$.
Thus, the attention weight will be the highest.
In addition, \cref{app-eq:rerope-angle-diff} increases when $t$ goes from $\midt_n$ to $\startt_n$ or $\endt_n$, leading to a decreased weight.

\heading{(iii)} \textit{The video token at the transition point between two events should attend equally to their text embeddings.}
\\
\heading{Proof}\ \ For $t = \startt_n$ or $\endt_n$, we always have the distance $\left \vert \tilde{t} - \tilde{t}^{\mathrm{mid}}_n \right \vert = L/2$.
Thus, the attention value is the same for video tokens at event borders.
This is only possible in ReRoPE as we rescale all events to have the same length.

\begin{figure}[!t]
    \centering
    \includegraphics[width=\linewidth]{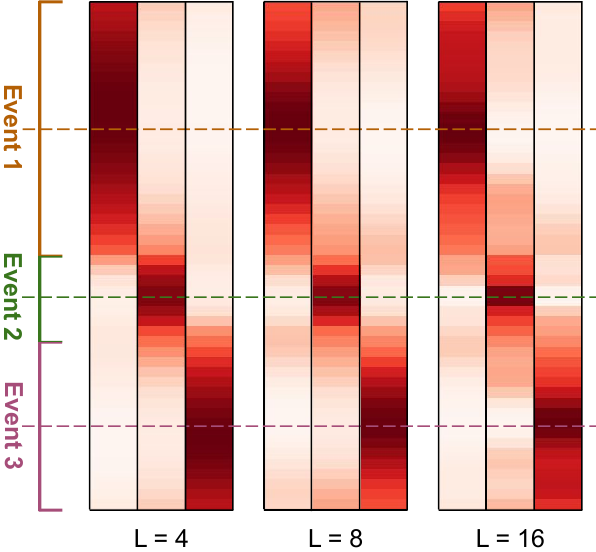}
    \vspace{\figcapmargin}
    \vspace{-4mm}
    \caption{
        \textbf{Comparison of ReRoPE with different rescaling length $L$.}
        We use the same random vector for video tokens and text embeddings to only visualize the bias introduced by positional encoding.
        We visualize the case where videos have a temporal dimension of 50, and there are three temporal captions.
    }
    \label{app-fig:rerope-vis-diff-L}
    \vspace{\figmargin}
    \vspace{1mm}
\end{figure}

\subsection{Visualizations of ReRoPE}
\label{app:rerope-vis-L}

In \cref{sec:ablation} of the main paper, we show that using different rescaling length $L$ in ReRoPE leads to similar results.
\cref{app-fig:rerope-vis-diff-L} visualizes the cross-attention map using $L=4$, $8$, and $16$.
The three attention maps are indeed similar, which explains why the performances are close.
We also notice that with a higher $L$, the attention map of each event becomes more concentrated.
It would be an interesting direction to study its effect in depth, which we leave for future work.

\begin{figure*}[t]
    \vspace{\pagetopmargin}
    \centering
    \begin{subfigure}{0.32\linewidth}
        \includegraphics[width=1.0\linewidth]{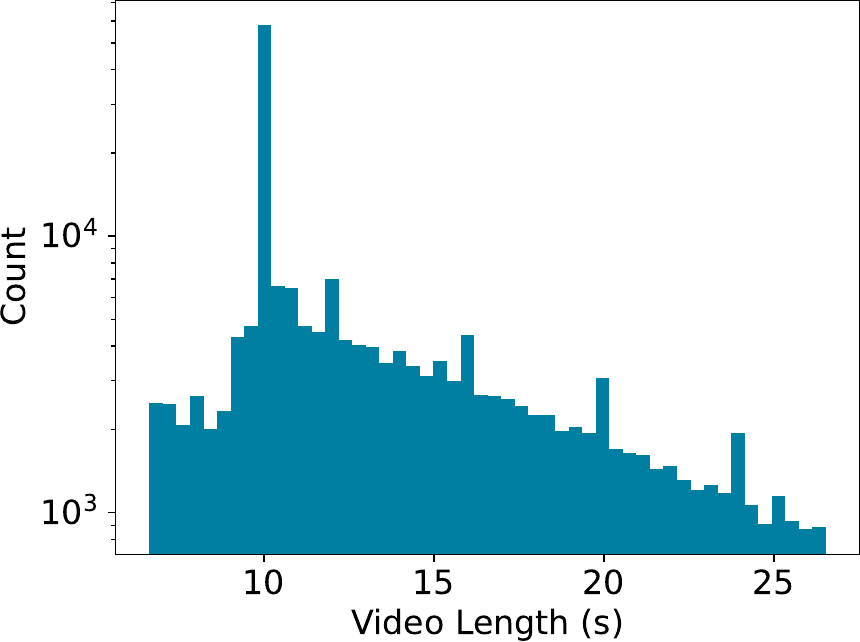}
        \vspace{-3mm}
    \end{subfigure}
    \hspace{1mm}
    \begin{subfigure}{0.32\linewidth}
        \includegraphics[width=1.0\linewidth]{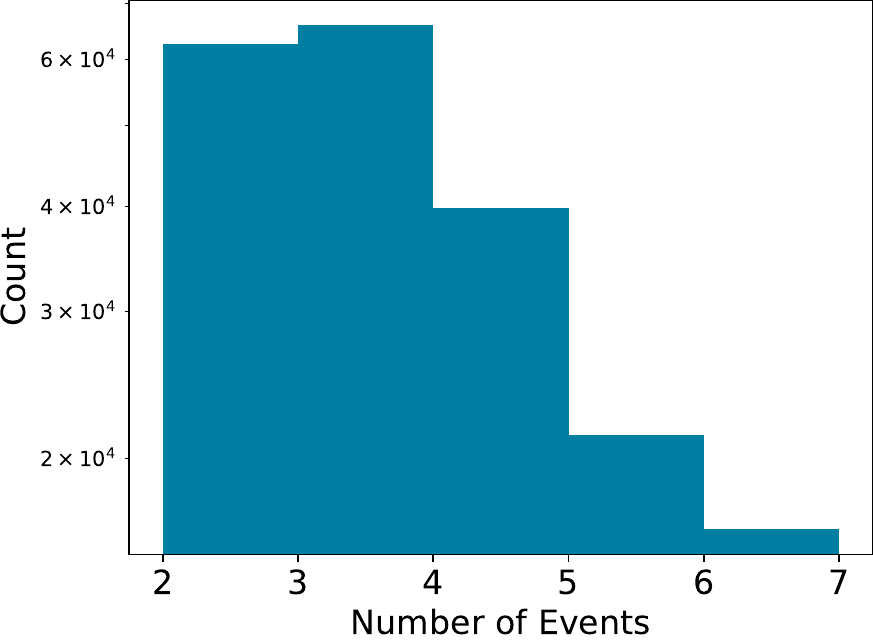}
        \vspace{-3mm}
    \end{subfigure}
    \hspace{1mm}
    \begin{subfigure}{0.32\linewidth}
        \includegraphics[width=1.0\linewidth]{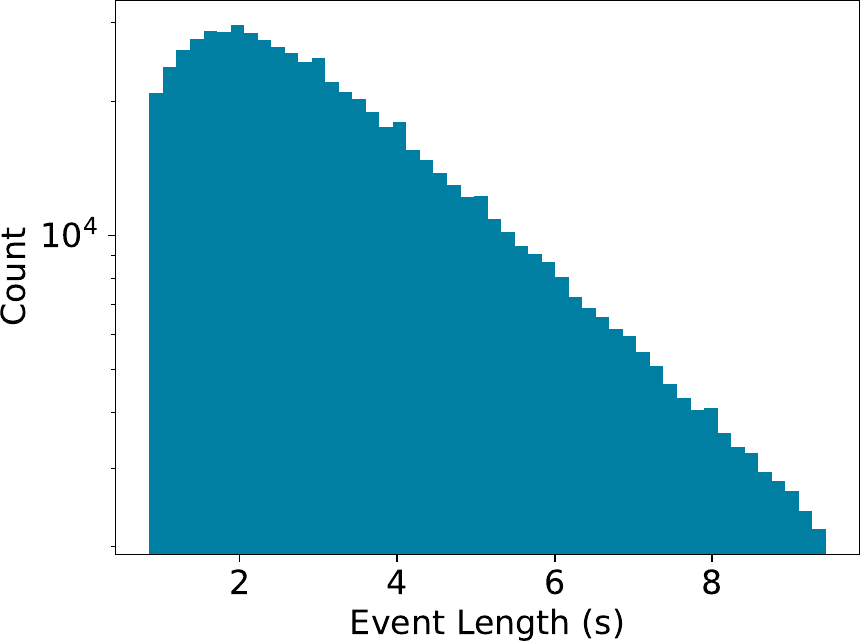}
        \vspace{-3mm}
    \end{subfigure}
    \vspace{-2mm}
    \caption{
        \textbf{Basic statistics of our training dataset.}
        We show the distribution of video length, the number of events per video, and the length of individual events.
        Most videos contain 2 to 4 events, and most events are under 5s.
    }
    \label{app-fig:dataset-stats}
    \vspace{\figmargin}
\end{figure*}

\section{Detailed Experimental Setup}
\label{app:more-exp-setup}

In this section, we provide full details on the datasets, baselines, evaluation settings, and the training and inference implementation details of our model.

\subsection{Training Data}
\label{app:training-data}

Before this work, there are mainly two types of video datasets that annotate open-set event captions and their precise timestamps.
One such field is dense video captioning~\cite{ActivityNetCaption,YouCook2,ViTTDenseCaptionDataset}.
However, these datasets are limited in scale (usually fewer than 10k videos), which makes it impossible to fine-tune a large-scale video generator.
Another field is video chaptering~\cite{VidChapters-7M}.
However, the temporal captions here are high-level chapter segmentation, where each annotated event is usually longer than one minute.
This is too long for current video diffusion models to be trained on.

Since our model requires large-scale and fine-grained video event annotations, we manually source videos from existing datasets~\cite{Panda-70M,HD-VILA} and annotate them, resulting in around 200k videos.
To condition the model on scene cuts, we run TransNetV2~\cite{TransNetV2} to detect scene boundaries on annotated videos with a confidence threshold of 0.5.

\cref{app-fig:dataset-stats} present some basic statistics of our dataset.
While our training videos have varying lengths, the number of events per video and the average event length are similar, which makes model training easier.

\heading{Data processing.}
The training dataset contains videos of different lengths, resolutions, and aspect ratios.
Following common practice~\cite{OpenSora,SDXL}, we use data bucketing, which groups videos into a fixed set of sizes.
Overall, we sample videos up to 512 resolution, and 10s during training.
We pad to or subsample 4 temporal captions for batch training.

\subsection{Evaluation Datasets}
\label{app:eval-data}

\heading{\holdout.}
We randomly sample 2k videos from our training data as a holdout testing set.
The prompts here are in-distribution with a minimum gap to training data.

\heading{StoryBench}~\cite{StoryBench} consists of videos collected from DiDeMo~\cite{DiDeMoDataset}, Oops~\cite{OopsDataset}, and UVO~\cite{UVODataset} datasets.
It annotates each video with a background caption and one or more temporal captions similar to our format.
We treat their background caption as the global caption in our setting, showing our model's generalization to out-of-distribution prompts.
We filter out videos with only a single event, leading to around 3k testing samples.

\heading{VBench}~\cite{VBench} is a comprehensive benchmark that tests different aspects of a video generation model.
It has 16 evaluation dimensions, each with a carefully collected list of text prompts.
Since we are interested in the dynamics of generated videos, we choose the \textit{Dynamic Degree} dimension, which provides 72 prompts.
Following the official evaluation protocol, we run each model to generate 5 videos using each prompt with 5 random seeds.

\subsection{Baselines}
\label{app:baselines}

We only compare to methods that can generate smoothly connected events and have released their code.

\heading{MEVG}~\cite{MEVG} is the state-of-the-art multi-event video generation method.
Given a sequence of event prompts, it generates the first video clip using the first event prompt.
Then, to generate the next event, it runs DDIM inversion~\cite{DDIM} to obtain the inverted noise latent of the previous clip, which is used to initialize the current noise latent.
Then, when denoising the current latent, it also introduces several losses to enforce latent at adjacent frames to be similar.
Original MEVG builds upon LVDM~\cite{LVDM} and VideoCrafter~\cite{VideoCrafter1} which are outdated.
For a fair comparison, we re-implement it based on our base model.
As far as we know, there is no prior work on inverting a rectified flow model, so we follow DDIM inversion to implement RF inversion which achieves similar results.
To handle both global and temporal captions, we generate the first clip by concatenating the global caption and the first temporal caption.
We keep other losses and hyper-parameters the same as in MEVG\footnote{MEVG did not release the code at the time of paper submission. We obtain the official code from authors through private email communication.}.

\heading{\imgcondmodel.}
We fine-tune our base model to support initial frame conditioned video generation.
The method is similar to MEVG, where we generate one event based on its own caption and the last frame of the previous clip.

\heading{\jointmodel} is a naive baseline that simply concatenates the global caption and all temporal captions to form a long prompt, and generates a video from it.

\noindent\textit{Remark.}
Since both MEVG and \imgcondmodel are autoregressive methods, they can only generate fixed-length videos for each event.
To enable comparison, we simply assume that the testing events all have the same duration when computing metrics.
For \jointmodel, it cannot separate the generation of different events.
We thus assume all events are uniformly distributed in the generated video.

\subsection{Evaluation Metrics}
\label{app:eval-metrics}

We identify three key aspects in multi-event text-to-video generation: visual quality, event text alignment, and event transition smoothness.
We report common metrics such as FID~\cite{FID}, FVD~\cite{FVD} for visual quality, and per-frame CLIP-score~\cite{CLIPScore,CLIP} for text alignment.
We have tried more advanced metrics such as X-CLIP-score~\cite{X-CLIP}, but found it to perform similarly as CLIP-score.

It is well-known that traditional automatic metrics are not aligned with human perceptions.
Recent works show that fine-tuning multi-modal LLMs on human feedback data can lead to more human-aligned video quality assessment metrics~\cite{VideoScore}.
We take the state-of-the-art method VideoScore which outputs five scores for a video.
We use the \textit{visual quality} and \textit{dynamic degree} output for visual quality, the \textit{text-to-video alignment} output for text alignment, and the \textit{temporal consistency} output for event transition smoothness.
We further run TransNetV2~\cite{TransNetV2} to compute the average number of cuts in generated videos to measure event transition smoothness.

For visual quality and event transition smoothness, we compute relevant metrics on the entire video.
We have also computed the visual quality of each event, and found it to be positively correlated with video-level results.
For text alignment, since we care about event generation, we take the start and end timestamps of each event, crop out a sub-clip from the generated video, and compute metrics between this sub-clip and the corresponding event prompt.

\begin{figure*}[!t]
    \vspace{\pagetopmargin}
    \centering
    \includegraphics[width=\linewidth]{imgs/qual/t2v/t2v-qual-ex2.pdf}
    \vspace{\figcapmargin}
    \vspace{-5mm}
    \caption{
        \textbf{Qualitative comparisons of T2V.}
    }
    \label{app-fig:more-t2v-qual-comparison}
    \vspace{\figmargin}
\end{figure*}

\begin{figure*}[!t]
    \vspace{3mm}
    \centering
    \includegraphics[width=\linewidth]{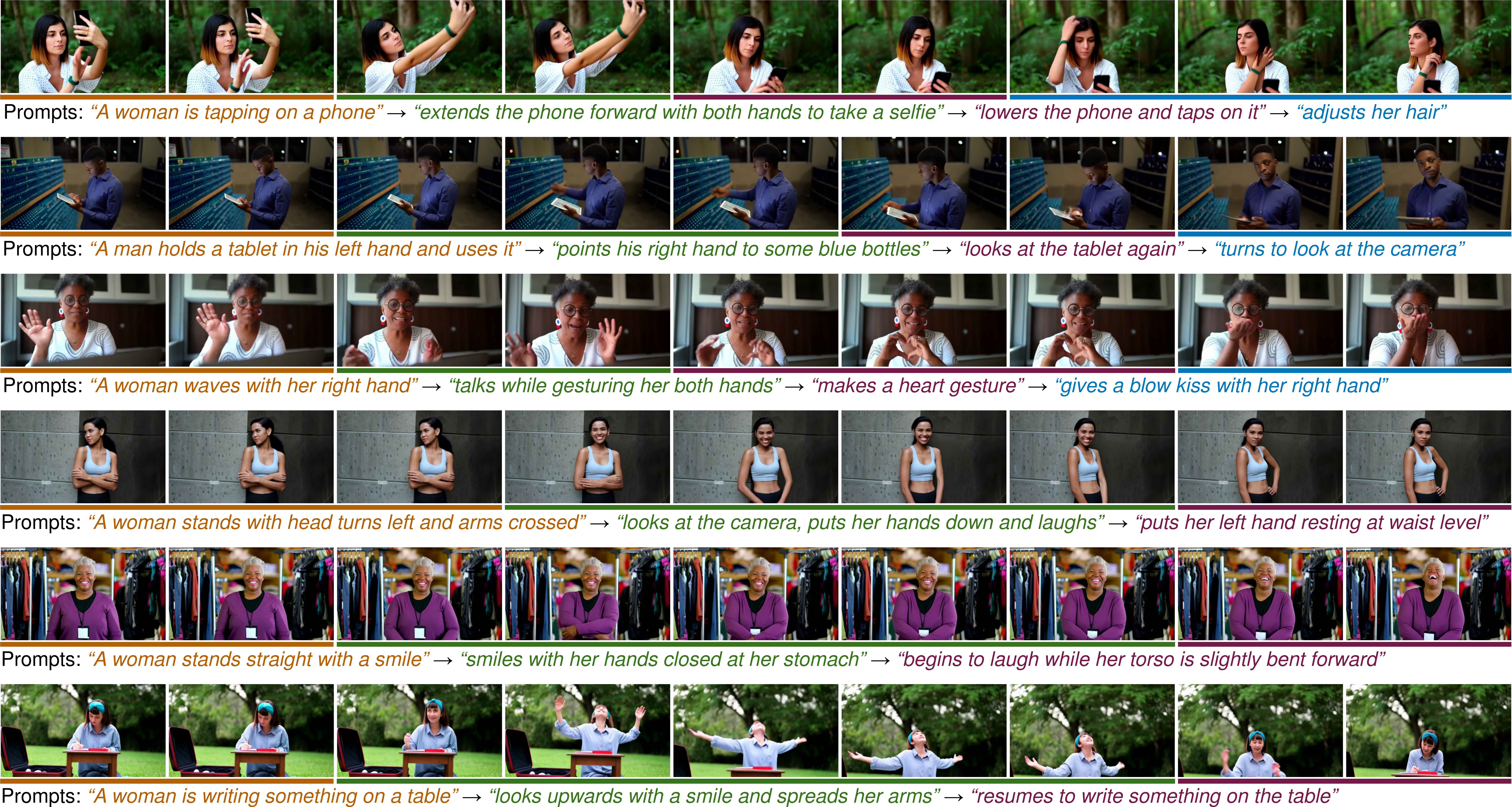}
    \vspace{\figcapmargin}
    \vspace{-4mm}
    \caption{
        \textbf{More T2V results from \algoNameFull.}
        Please see our \href{https://mint-video.github.io/\#more-our-results}{project page} for more results.
    }
    \label{app-fig:more-t2v-qual-ours}
    \vspace{\figmargin}
\end{figure*}

\subsection{Implementation Details}
\label{app:out-impl-details}

\heading{Base model.}
Our base text-to-video generator adopts the latent Diffusion Transformer framework~\cite{DiT}.
It leverages a MAGVIT-v2~\cite{MAGVITv2} as the autoencoder and a deep cascade of DiT blocks as the denoising backbone.
The autoencoder is similar to the one in CogVideoX~\cite{CogVideoX}, which downsamples the spatial dimensions by 8$\times$ and the temporal dimension by 4$\times$.
Our backbone has 32 DiT blocks.
Each block is similar to the one in Open-Sora~\cite{OpenSoraPlan}, which consists of a 3D self-attention layer running on all video tokens, a cross-attention layer between video tokens and T5 text embeddings~\cite{T5} of the input prompt, and an MLP.
We do not use absolute positional encoding on video tokens.
Instead, we apply RoPE in self-attention, which is factorized into spatial and temporal axes, similar to \cite{OpenSoraPlan}.
Finally, we use FlashAttention~\cite{FlashAttention} in both self-attention and cross-attention.
\\
The base model adopts the rectified flow training objective~\cite{FlowMatching,RFSampler}.
We follow Stable Diffusion 3~\cite{SD3} to choose the sampling parameters for the diffusion process.

\heading{\algoNameFull model.}
We fine-tune \algoNameFull from the base model to enable temporal caption control.
We copy weights from the original cross-attention layer to initialize our added temporal cross-attention layer to accelerate convergence, since both layers take in the same text modality.
Following prior works~\cite{GLIGEN}, we introduce a scaling factor that is initialized as 0, and we pass it through a $\mathrm{Tanh(\cdot)}$ activation to multiply with the temporal cross-attention layer output.
Such a design has been shown to stabilize model training.

\heading{Training.}
We use AdamW~\cite{AdamW} to fine-tune the entire model with a batch size of 512 for 15k steps.
We use a low learning rate of $1 \times 10^{-5}$ for the pre-trained weights, and a higher one of $1 \times 10^{-4}$ for the added weights.
Both learning rates are linearly warmed up in the first 1k steps and stay constant.
A gradient clipping of 0.05 is applied to stabilize training.
To apply classifier-free guidance (CFG)~\cite{CFG}, we randomly drop the text embedding of global and temporal captions (i.e., setting them as zeros) with a probability of 10\%.
Notice that when dropping the temporal captions, we drop all events together and also set the event timestamps to zeros.
We implement our model using PyTorch~\cite{PyTorch} and conduct training on NVIDIA A100 GPUs.

\heading{Inference.}
We use the rectified flow sampler~\cite{RFSampler} with 256 sampling steps and a classifier-free guidance~\cite{CFG} scale of 8 to generate videos.
We also use interval guidance~\cite{CFGInterval} in CFG to mitigate the oversaturation issue, which only applies CFG between [25, 100] sampling steps.
We have tried using separate CFG for global and temporal captions similar to in \cite{InstructPix2Pix}, but did not find it to improve results.

\begin{figure*}[!t]
    \vspace{\pagetopmargin}
    \centering
    \includegraphics[width=\linewidth]{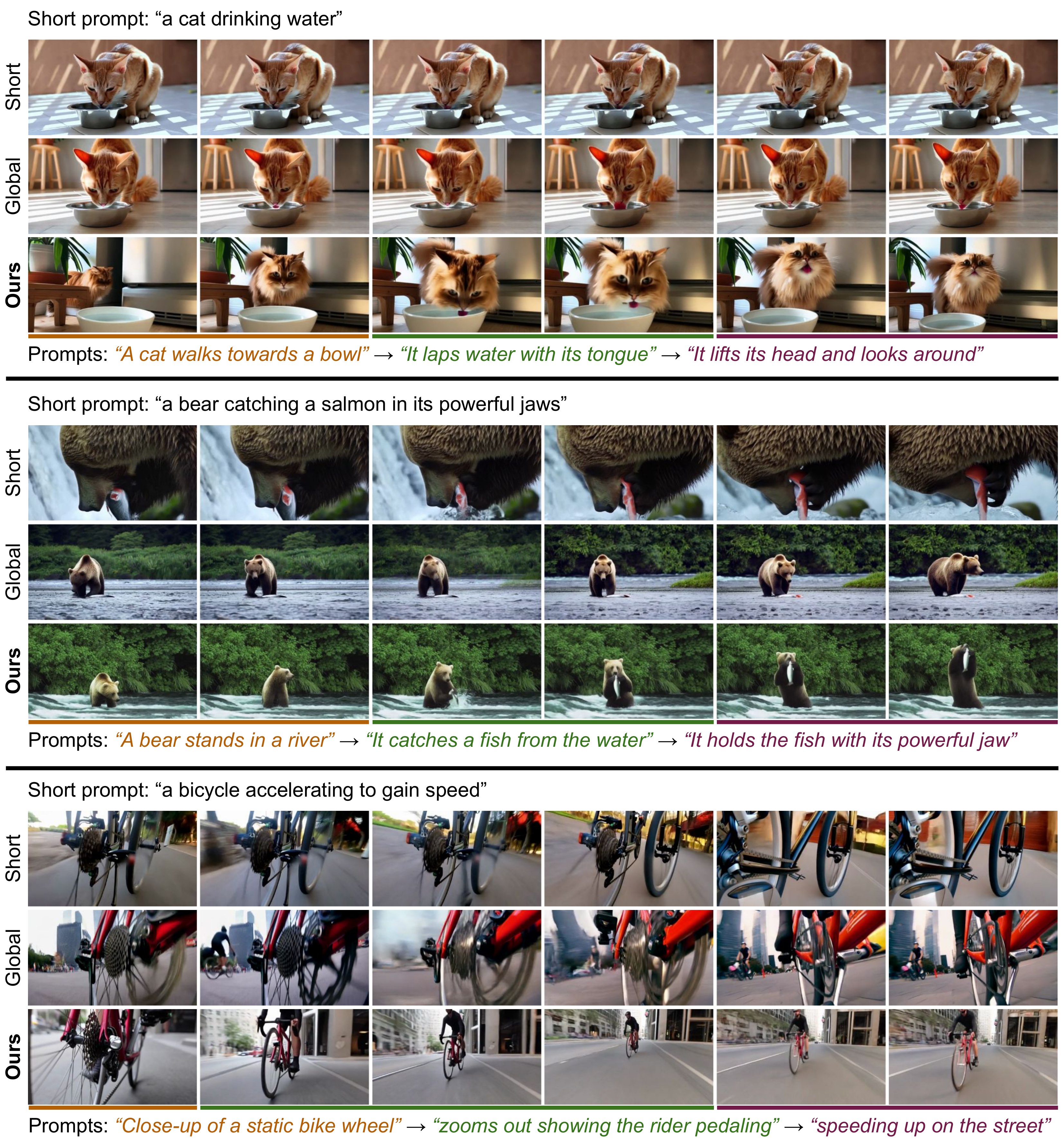}
    \vspace{\figcapmargin}
    \vspace{-4mm}
    \caption{
        \textbf{Prompt enhancement results on VBench.}
        We can generate more interesting videos from a simple prompt.
        This highlights the flexible dynamics control ability brought by the temporal captions.
        Please see our \href{https://mint-video.github.io/\#prompt-enhancement}{project page} for video results.
    }
    \label{app-fig:vbench-enhance-qual-results}
    \vspace{\figmargin}
    \vspace{1mm}
\end{figure*}

\section{More Results}
\label{app:more-results}

\begin{figure*}[!t]
    \vspace{\pagetopmargin}
    \centering
    \includegraphics[width=\linewidth]{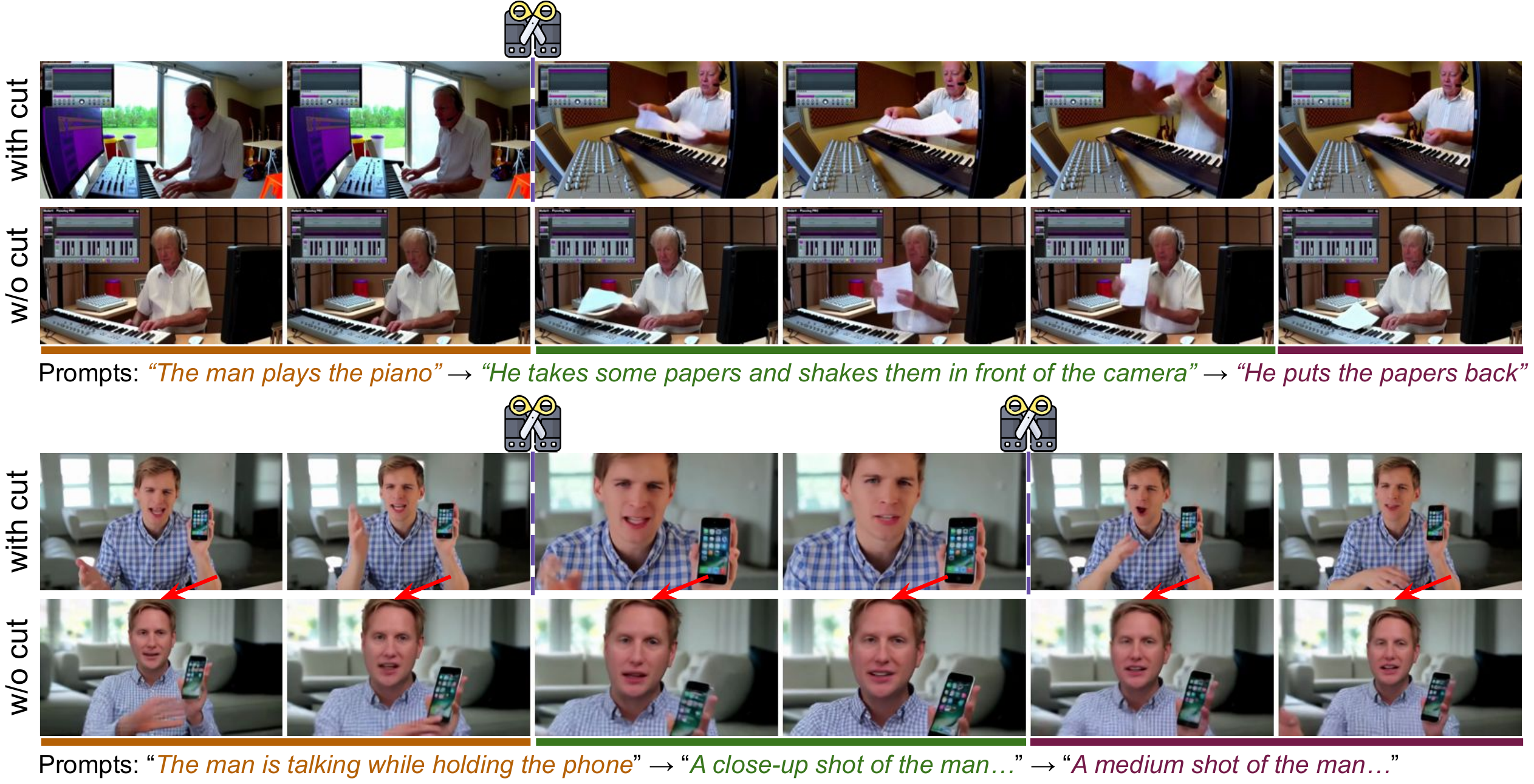}
    \vspace{\figcapmargin}
    \vspace{-4mm}
    \caption{
        \textbf{Generated videos with and without scene cut input.}
        In each example, the first row is generated by inputting the scene cut at the illustrated timestamps, while the second row is by zeroing out the scene cut input.
        When using the scene cut, the model is able to generate a shot transition at desired timestamps, while keeping the subject consistent.
        In the second example, the model generates smooth zoom-in and zoom-out effects when zeroing out scene cuts.
        Please see our \href{https://mint-video.github.io/\#scene-cut-condition}{project page} for more results.
    }
    \label{app-fig:scenecut-control-qual-results}
    \vspace{\figmargin}
    \vspace{1mm}
\end{figure*}

\subsection{More Qualitative Results on T2V}
\label{app:more-qual-comparison}

\cref{app-fig:more-t2v-qual-comparison} presents more qualitative comparisons with baselines.
\jointmodel only generates the woman writing on a paper while ignoring the subsequent events.
\imgcondmodel is able to synthesize a smooth transition between the first and the second event, but it fails to generate the third event.
This is because conditioning on generated frames leads to video stagnation and results in frozen frames.
MEVG generates each event well, but they are connected with abrupt shot transitions and completely different subjects.
This is due to the free-form event captions we use, which change the subjects frequently.
As a result, the inversion technique in MEVG cannot preserve the subjects well.
So far, there is no inversion method designed for rectified flow models.
Overall, \algoNameFull is the only method that successfully generates all events with smooth transitions and consistent entities.

We show more qualitative results of \algoNameFull in \cref{app-fig:more-t2v-qual-ours}.
Human-related subjects are known to be challenging in visual generation tasks.
Yet, the results demonstrate our flexible control of human action sequences and time lengths.

\subsection{Prompt Enhancement}
\label{app:prompt-enhancement-vbench}

Our prompt enhancer is built upon GPT-4~\cite{GPT-4} and can extend a short prompt to a detailed global caption and multiple temporal captions with reasonable event timestamps.
We provide the instruction we used on our \href{https://mint-video.github.io/src/vbench/prompt.txt}{project page}.
It is inspired by recent works~\cite{MEVG,VideoStudio} and uses in-context examples from our dataset for better performance.

We show more prompt enhancement results using VBench prompts in \cref{app-fig:vbench-enhance-qual-results}.
Thanks to the powerful LLM, our prompt enhancer can extend a short prompt to reasonable sequential events, covering rich object motion and camera movement.
\algoNameFull can then generate more interesting and ``eventful" videos from the extended prompt.
This highlights the unique capability of our method, opening up a new direction towards more user-friendly video generation.

\subsection{Scene Cut Conditioning}
\label{app:scene-cut-conditioning}

As shown in the ablation, removing scene cut conditioning leads to undesired shot transitions in generated videos.
A closer inspection reveals that the generation of cuts is sensitive to the text prompt of an event.
When it contains a description of the camera shot (\eg ``a close-up view of"), it is more likely to introduce a cut.
In contrast, explicitly conditioning on scene cuts frees us from this issue.

We show some qualitative scene cut control results in \cref{app-fig:scenecut-control-qual-results}.
\algoNameFull is able to generate shot transitions at desired timestamps, while preserving subject identities.
When zeroing out the scene cut input, we can get cut-free videos which validates our design.
Finally, we show that our model can switch between sudden camera changes or gradual zoom-in and zoom-out effects, enabling fine-grained control.

An interesting direction is to learn different types of scene transitions such as jump cut, dissolve, and wipe.
Since our goal is to retain training data instead of learning fancy transition control, we leave this for future work.

\begin{figure*}[!t]
    \vspace{\pagetopmargin}
    \centering
    \includegraphics[width=\linewidth]{imgs/qual/t2v/time-control-qual-ex1.pdf}
    \vspace{\figcapmargin}
    \vspace{-4mm}
    \caption{
        \textbf{Generated videos with different event time spans.}
        In each example, we offset the start and end timestamps of all events by a specific number of seconds.
        Results show that \algoNameFull enables fine-grained event timing control while keeping the subjects' appearances to be roughly the same.
        This capability is very useful for controllable video generation.
        Please see our \href{https://mint-video.github.io/\#time-control}{project page} for more results.
    }
    \label{app-fig:time-control-qual-results}
    \vspace{\figmargin}
    \vspace{1mm}
\end{figure*}

\subsection{Event Time Span Control}
\label{app:event-time-control}

\algoNameFull supports fine-grained control of event time span.
To show this, we take a sample from our dataset and offset the start and end timestamps of all events by a specific value.
\cref{app-fig:time-control-qual-results} presents the results, where each video generates events following its new timing.
In addition, we can roughly keep the appearance of the main subject and background unchanged.
\algoNameFull is the first video generator in the literature that achieves this control ability.
We view it as an important step towards a practical content generation tool.

\begin{table}[t]
    \vspace{1mm}
    \centering
    \setlength{\tabcolsep}{12pt}
    \footnotesize
    \begin{tabular}{l|ccc}
        \toprule
        \textbf{Method} & FID $\downarrow$ & FVD $\downarrow$ & CLIP-score $\uparrow$ \\
        \midrule
        \multicolumn{4}{c}{\textit{Task: T2V (a.k.a. story generation in \cite{StoryBench})}} \\
        \midrule
        Phenaki & 273.41 & 998.19 & 0.210 \\
        \textbf{Ours} & \textbf{40.87} & \textbf{484.44} & \textbf{0.284} \\
        \midrule
        \multicolumn{4}{c}{\textit{Task: I2V (a.k.a. story continuation in \cite{StoryBench})}} \\
        \midrule
        Phenaki & 240.21 & 674.5 & 0.219 \\
        \textbf{Ours} & \textbf{21.85} & \textbf{314.59} & \textbf{0.273} \\
        \bottomrule
    \end{tabular}
    \vspace{\tablecapmargin}
    \caption{
        \textbf{Comparison with Phenaki on StoryBench.}
        We compare with the zero-shot variant Phenaki-Gen-ZS in their paper~\cite{StoryBench} since our model is not fine-tuned on StoryBench.
        We clearly outperform Phenaki across all metrics in both tasks.
    }
    \label{app-tab:storybench-quan-results-vs-phenaki}
    \vspace{\tablemargin}
\end{table}

\subsection{StoryBench Comparison with Phenaki}
\label{app:storybench-compare-phenaki}

The original StoryBench paper~\cite{StoryBench} proposed a baseline for their dataset, which runs Phenaki~\cite{Phenaki} to generate events in an autoregressive way.
However, they conducted evaluation on a much lower resolution (160$\times$96), and neither their code nor pre-trained weights were released, making a direct comparison hard.
We still compare with them in \cref{app-tab:storybench-quan-results-vs-phenaki} for completeness.
We only report metrics that both papers evaluate, which cover visual quality (FID, FVD) and text alignment (CLIP-score).
\algoNameFull significantly outperforms Phenaki across all metrics in both T2V and I2V tasks.
This demonstrates the effectiveness of fine-tuning from a large-scale pre-trained video model.

\subsection{Comparison with SOTA Video Generators}
\label{app:more-compare-with-sota}

To show that sequential event generation is a common failure case of even SOTA video generators, we present more results in \cref{app-fig:compare-with-sota-models-1} and \cref{app-fig:compare-with-sota-models-2}.
One surprising observation we had is that, when using prompts following the official guideline of these models (\eg using the LLM provided by CogVideoX to enhance prompts), the model only generates the first event and ignores all subsequent ones.
Only if we directly concatenate event captions without specifying global properties such as camera motion, background description, and detailed subject attributes (\ie directly use prompts like ``A person first do A, then do B, and finally do C"), the model starts to generate some events transitions.\footnote{The detailed prompt does not exceed the maximum input text length of these models, so context length is not the reason here.}
One possible cause is that in the training data of these models, videos with sequential events are never annotated with such detailed global properties.
However, since we do not have access to their training details, we can't figure out the true reason behind it.
Therefore, we just use naively concatenated prompts to generate all results.
The prompts we used for these models can be found on our \href{https://mint-video.github.io/src/vs_sota/prompts.txt}{project page}.
Notably, this workaround prevents us from using detailed captions to control the scene and subjects, which greatly affects the controllability of these models.

Still, when prompted with a text that contains multiple events, these models have three common failure modes:
\begin{itemize}
    \item[1.] Only generates partial events, and completely ignores the remaining ones. For example, in the third example in \cref{app-fig:compare-with-sota-models-1}, all models miss the ``blow kiss" action;
    \item[2.] Generates events in the wrong order or ``merge" multiple events. For example, in the last example in \cref{app-fig:compare-with-sota-models-1}, Kling 1.5 generates the man with his hand under his mouth at the beginning of the video. Yet, this should happen last;
    \item[3.] Bind wrong actions or properties to subjects. For example, in the first example in \cref{app-fig:compare-with-sota-models-2}, Gen-3 Alpha generates a woman coming into the frame instead of a man.
\end{itemize}

\noindent\textit{Remark.}
There might be other ways to fix this issue without using temporally-grounded captions as in \algoNameFull.
For example, one may fine-tune the model on video datasets annotated with detailed sequential event information~\cite{LVD-2M-Temporal-Dense-Caption-Dataset}.
Still, this will not allow precise control over the start and end times of events, which is a unique capability of our model.

\begin{figure*}[!t]
    \vspace{\pagetopmargin}
    \centering
    \includegraphics[width=\linewidth]{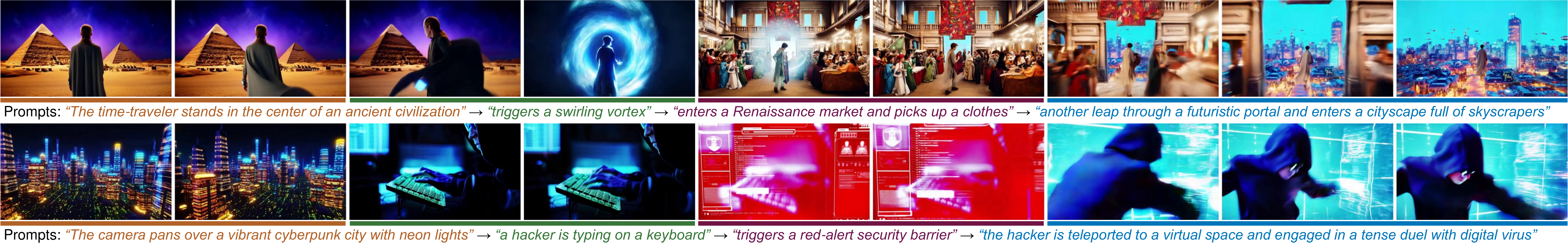}
    \vspace{\figcapmargin}
    \vspace{-4mm}
    \caption{
        \textbf{Generated videos with extreme dynamics.}
        We prompt \algoNameFull to generate scene cuts at event boundaries, leading to explicit scene changes and large dynamics.
    }
    \label{app-fig:very-dynamic-results}
    \vspace{\figmargin}
    \vspace{2mm}
\end{figure*}

\begin{figure*}[!t]
    \centering
    \includegraphics[width=\linewidth]{imgs/qual/ood-results.pdf}
    \vspace{\figcapmargin}
    \vspace{-4mm}
    \caption{
        \textbf{Generated videos with out-of-distribution prompts.}
        After fine-tuning, \algoNameFull still possesses the base model's ability to generate novel concepts.
        Please see our \href{https://mint-video.github.io/\#ood-results}{project page} for more results.
    }
    \label{app-fig:ood-results}
    \vspace{\figmargin}
    \vspace{1mm}
\end{figure*}

\heading{Quantitative comparison with Kling.}
Running Kling on all test prompts will incur unreasonable API costs ($\sim$\$5k).
Therefore, we ran it on the 200 prompts used in our user study, and conducted a user study with 20 participants per prompt similar to our main experiment.
Due to a weaker base model, \algoNameFull achieves a lower Visual Quality (31.55\% win rate).
Nevertheless, \algoNameFull clearly outperforms Kling in all three event-related metrics (73.18\% in Text Alignment, 69.93\% in Event Timing, and 68.27\% in Event Transition).

\subsection{Generating Videos with Extreme Dynamics}

We prompt \algoNameFull to generate videos with extreme dynamics.
Thanks to the scene cut conditioning, we enable explicit scene changes in generated videos as shown in \cref{app-fig:very-dynamic-results}.

\subsection{Out-of-Distribution Prompts}
\label{app:ood-prompts}

\algoNameFull is fine-tuned on temporal caption videos that mostly describe human-centric events.
In the paper, we have shown some non-human results such as animals and traffics.
Here, we show that our model still possesses the ability to generate novel concepts and their combinations, which is an important property of large-scale pre-trained video generators.
As shown in \cref{app-fig:ood-results}, \algoNameFull generates out-of-distribution characters such as warriors and astronaut, scenes such as starships in the space, and non-existing events such as a cat doing yoga.
This proves that our model does not forget the rich pre-training knowledge in the base model.

\section{Limitations and Future Works}
\label{app:failure-case}

\algoNameFull is fine-tuned from a pre-trained text-to-video diffusion model, and thus we are bounded by the capacity of the base model.
For example, it is challenging to generate human hands or scenes involving complex physics.

When generating an event involving multiple subjects, \algoNameFull may fail to associate attributes and actions to the correct subject.
Similar to the \textit{temporal binding} problem we try to address in this paper, we believe this issue can be solved with \textit{spatial binding}.
For example, by grounding subjects with bounding boxes and attribute labels~\cite{GLIGEN,LLMGroundedDM,Boximator}.

Finally, \algoNameFull sometimes fails to associate entities specified in the global caption and temporal captions.
Such association requires complex reasoning of the text conditioning, and may be resolved by simply scaling up the training data.

Please refer to our \href{https://mint-video.github.io/\#failure-case}{project page} for video examples and detailed analysis of these failure cases.

\heading{Future works.}
It is interesting to enhance our model with recent progress in training-free long video generation techniques~\cite{Gen-L-Video,StreamingT2V,FreeNoise}.
Another direction is to combine \algoNameFull with video personalization methods~\cite{VideoBooth,VideoDirectorGPT,VideoStudio,VideoAlchemist} to enable both fine-grained control within a shot and subject consistency across shots for minute-long video creation.

\clearpage
\thispagestyle{empty}
\begin{figure*}[!t]
    \vspace{\pagetopmargin}
    \vspace{-2mm}
    \centering
    \includegraphics[width=\linewidth]{imgs/qual/t2v/t2v-qual-vs-sota-ex1-w-mochi.pdf}
    \vspace{\figcapmargin}
    \vspace{-4mm}
    \caption{
        \textbf{More comparisons with SOTA video generators.}
        We run SOTA open-source models CogVideoX~\cite{CogVideoX} and Mochi~\cite{Mochi}, and commercial models Kling 1.5~\cite{KLING1_5} and Gen-3 Alpha~\cite{Gen3Alpha} using their online APIs.
        Please see our \href{https://mint-video.github.io/\#compare-with-sota}{project page} for video results.
    }
    \label{app-fig:compare-with-sota-models-1}
    \vspace{\figmargin}
    \vspace{1mm}
\end{figure*}

\clearpage
\thispagestyle{empty}
\begin{figure*}[!t]
    \vspace{\pagetopmargin}
    \centering
    \includegraphics[width=\linewidth]{imgs/qual/t2v/t2v-qual-vs-sota-ex2-w-mochi.pdf}
    \vspace{\figcapmargin}
    \vspace{-4mm}
    \caption{
        \textbf{More comparisons with SOTA video generators.}
        We run SOTA open-source models CogVideoX~\cite{CogVideoX} and Mochi~\cite{Mochi}, and commercial models Kling 1.5~\cite{KLING1_5} and Gen-3 Alpha~\cite{Gen3Alpha} using their online APIs.
        Please see our \href{https://mint-video.github.io/\#compare-with-sota}{project page} for video results.
    }
    \label{app-fig:compare-with-sota-models-2}
    \vspace{\figmargin}
    \vspace{1mm}
\end{figure*}

\end{document}